\documentclass[10pt]{article} % For LaTeX2e

\usepackage[accepted]{rlj} % Should be uncommented for the camera-ready
\usepackage{amsmath}
\usepackage{amssymb}            % Defines common symbols like \mathbb R
\usepackage{mathtools}          % Extends amsmath, providing common math tools
\usepackage{amsthm}
\usepackage{mathrsfs}           % Enables \mathscr, which can work in cases that \mathcal does not
\usepackage{graphicx}           % For including images
\usepackage{subcaption}         % Allows for the use of subfigures and subcaptions
\usepackage[space]{grffile}     % For spaces in image names
\usepackage{url}                % For displaying URLs
\usepackage{lipsum}             % For placeholder text
\usepackage{bm}
\usepackage{hyperref}

\theoremstyle{plain}

\theoremstyle{definition}

\theoremstyle{remark}

\definecolor{codegray}{gray}{0.9}

\newcommand{\repoURL}{\url{https://github.com/taodav/pobax}}
\title{Benchmarking Partial Observability in Reinforcement Learning with a Suite of Memory-Improvable Domains}

\setrunningtitle{Benchmarking Partial Observability in RL with a Suite of Memory-Improvable Domains}

\author{
Ruo Yu Tao \textsuperscript{1,$\dagger$}, Kaicheng Guo \textsuperscript{1, $\dagger$}, Cameron Allen \textsuperscript{2}, George Konidaris \textsuperscript{1}
}

\emails{
\{ruoyutao,kaicheng\_guo\}@brown.edu
}

\affiliations{
$^{1}$\textbf{Brown University}
$^{2}$\textbf{UC Berkeley}\\

\par % If including additional comments like below, use \par to add some whitespace. 
$^\dagger$ Equal contribution.
}

\contribution{
    We investigate the efficacy of partially observable benchmarks in measuring an algorithm's ability to mitigate partial observability.
    }
    {
    None
    }

\contribution{
    We introduce the memory improvability property: a partially observable benchmark is memory improvable if there is a gap between agents with more or less state information, all other factors roughly equal.
    }
    {
    None
    }

\contribution{
    We categorize popular forms of partial observability, and present a list of representative environments that covers these categories.
    }
    {
    This categorization does not cover all forms of partial observability.
    }

\contribution{
    We present the open-source POBAX benchmark: a suite of memory improvable environments designed to test an algorithm's ability to mitigate partial observability. POBAX is entirely implemented in JAX, allowing for fast and GPU-scalable experimentation. 
    }
    {
    While previous benchmarks exist for partial observability~\citep{rajan2021mdp, morad2023popgym, osband2020bsuite}, these works do not cover such breadth of environments.
    }

 % Your keywords

\summary{
Mitigating partial observability is a necessary but challenging task for general reinforcement learning algorithms.
To improve an algorithm's ability to mitigate partial observability, researchers need comprehensive benchmarks to gauge progress.
Most algorithms tackling partial observability are only evaluated on benchmarks with simple forms of state aliasing, such as feature masking and Gaussian noise. 
Such benchmarks do not represent the many forms of partial observability seen in real domains, like visual occlusion or unknown opponent intent.
We argue that a partially observable benchmark should have two key properties. The first is coverage in its forms of partial observability, to ensure an algorithm's generalizability.
The second is a large gap between the performance of a agents with more or less state information, all other factors roughly equal.
This gap implies that an environment is memory improvable: where performance gains in a domain are from an algorithm's ability to cope with partial observability as opposed to other factors.
We introduce best-practice guidelines for empirically benchmarking reinforcement learning under partial observability, as well as the open-source library POBAX: Partially Observable Benchmarks in JAX.
We characterize the types of partial observability present in various environments and select representative environments for our benchmark. These environments include localization and mapping, visual control, games, and more.
Additionally, we show that these tasks are all memory improvable and require hard-to-learn memory functions, providing a concrete signal for partial observability research.
This framework includes recommended hyperparameters as well as algorithm implementations for fast, out-of-the-box evaluation, as well as highly performant environments implemented in JAX for GPU-scalable experimentation. 
}

\begin{document}

\makeCover  % Create the cover page
\maketitle  % Make the title section

\begin{abstract}

Mitigating partial observability is a necessary but challenging task for general reinforcement learning algorithms.
To improve an algorithm's ability to mitigate partial observability, researchers need comprehensive benchmarks to gauge progress.
Most algorithms tackling partial observability are only evaluated on benchmarks with simple forms of state aliasing, such as feature masking and Gaussian noise. 
Such benchmarks do not represent the many forms of partial observability seen in real domains, like visual occlusion or unknown opponent intent.
We argue that a partially observable benchmark should have two key properties. The first is coverage in its forms of partial observability, to ensure an algorithm's generalizability.
The second is a large gap between the performance of a agents with more or less state information, all other factors roughly equal.
This gap implies that an environment is memory improvable: where performance gains in a domain are from an algorithm's ability to cope with partial observability as opposed to other factors.
We introduce best-practice guidelines for empirically benchmarking reinforcement learning under partial observability, as well as the open-source library POBAX: Partially Observable Benchmarks in JAX.
We characterize the types of partial observability present in various environments and select representative environments for our benchmark. These environments include localization and mapping, visual control, games, and more.
Additionally, we show that these tasks are all memory improvable and require hard-to-learn memory functions, providing a concrete signal for partial observability research.
This framework includes recommended hyperparameters as well as algorithm implementations for fast, out-of-the-box evaluation, as well as highly performant environments implemented in JAX for GPU-scalable experimentation.

\end{abstract}

%%%%%%%%%%%%%%%%%%%%%%%%%%%%%%%%%%%%%%%%%%%%%%%%%%%%%%%%%%%%%%%%
%% Section: Submission of papers to RLJ/RLC
%%%%%%%%%%%%%%%%%%%%%%%%%%%%%%%%%%%%%%%%%%%%%%%%%%%%%%%%%%%%%%%%
\section{Introduction}
\label{sec:intro}

Reinforcement learning~\citep{sutton2018book} algorithms are being deployed to increasingly complex domains where \textit{partial observability}~\citep{Kaelbling98} is a fundamental problem. A system is partially observable if its observations contain only partial information about the underlying state.
In this setting, agents cannot make effective decisions without reasoning about their past. 
Resolving partial observability is a necessary but typically challenging task~\citep{zhang2012covering}, and many system designers try to circumvent this issue with hand-designed environment-specific features~\citep{mnih15dqn, Bellemare2020balloon}.
The human engineering effort required to resolve partial observability environment by environment reveals the crux of the problem: there exist many different forms of partial observability, each with their own challenges.

To tackle partial observability, researchers develop history summarization algorithms through testing on benchmark partially observable tasks. The classic T-Maze~\citep{bakker2001reinforcement} problem was used to test long-term recall with LSTMs~\citep{hochreiter1997lstm} in reinforcement learning. 
The RockSample~\citep{smith2004rocksample} task was originally used to develop partially observable planning algorithms and their capabilities on large state spaces.

Current benchmarks are narrow in their scope of state aliasing, bringing into question whether performance on the benchmark translates to other forms of partial observability.
The best-known example is the Atari benchmark~\citep{bellemare13arcade}, where using only a single frame is partially observable~\citep{hausknecht15drqn}. Similarly, masked continuous control~\citep{han2020variational} is a popular benchmark where velocity or positional state information is hidden. Half of the masked continuous control tasks, the agent only requires a few previous time steps to gauge velocity information to recover a Markov state.
These benchmarks represent a narrow sampling of partial observability, but constitute a substantial fraction of empirical evaluations~\citep{Ni2021RecurrentMR, ni2023when, zhao2023odebased,lu2024rethinking}. 
Although other benchmarks test on more forms of state aliasing~\citep{morad2023popgym, beattie2016dmlab},
individual benchmarks lack coverage across the categories of partial observability and often lack justification as to why the selected tasks are good benchmark tasks. In some cases, performance on a partially observable benchmark depends more on implementation details rather than an algorithm's ability to mitigate partial observability~\citep{Ni2021RecurrentMR}.

Beyond good coverage of the forms of partial observability, a useful benchmark must have a clear signal for evaluating an algorithm's ability to mitigate partial observability. We argue that one such valuable signal is \textit{memory improvability}.
An environment is \textit{memory improvable} if a gap exists between the performance of agents imbued with more or less state information. This implies that using memory to mitigate partial observability will improve performance in this environment. 
The performance gap between observations that are partial and those that are (more) complete is exactly the gap that an agent mitigating partial observability ought to close.
A large gap indicates that a particular environment can benefit from adding memory; a small or non-existent gap indicates that either the partial observability is not a major issue, or there is some other confounding factor---e.g.
featurization scheme, learning dynamics or hyperparameters.

We introduce a new open-source benchmark, POBAX\footnote{Code: \repoURL}: Partially Observable Benchmarks in JAX. 
Since testing on all forms of partial observability is untenable, we categorize the different forms of partial observability and select representative environments for our benchmark to ensure that we have coverage of the space of task types.
POBAX is a comprehensive suite of new and existing partially observable environments that cover all state aliasing categories of interest described here. These environments include tasks such as localization and mapping, visual control, games and more. 
Besides requiring hard-to-learn memory, these environments are all memory improvable; as we add more information into the state representation, we see an increase in performance. To show the utility of our benchmark, we test three popular reinforcement learning algorithms designed for mitigating partial observability.
We also recommend per-environment hyperparamters for out-of-the-box evaluation of memory learning algorithms.
The benchmark is also entirely implemented in JAX~\citep{jax2018github}, allowing for fast simulation and GPU-scalable experiments. 

\section{Background and Related Work}
\label{sec:background}

We use Markov decision processes (MDPs)~\citep{Puterman94} and their extension, partially observable Markov decision processes (POMDPs)~\citep{Kaelbling98} as the framework for sequential decision making in an unknown environment. An MDP consists of a state space $\mathcal{S}$, action space $\mathcal{A}$, reward function $R: \mathcal{S} \times \mathcal{A} \rightarrow \mathbb{R}$, stochastic transition function $T: \mathcal{S} \times \mathcal{A} \rightarrow \Delta \mathcal{S}$, initial state distribution $p_0 \in \mathcal{S}$, and discount factor $\gamma \in [0, 1]$. The goal of an agent interacting with an MDP is to learn a policy $\pi_{\mathcal{S}}: \mathcal{S} \rightarrow \Delta \mathcal{A}$ which tries to maximize its expected discounted returns $V_{\pi_{\mathcal{S}}}(s) = \mathbb{E}_{\pi_{\mathcal{S}}}\left[ \sum_{i = 0}^{\infty}\gamma^i R_{t + i} \right]$.
In the POMDP framework, an agent receives observations $o \in \Omega$ through an observation function $\Phi: \mathcal{S} \rightarrow \Delta \Omega$ that maps the underlying hidden states to potentially incomplete state observations. These observations no longer have the Markov property: the observation $o_t$ and action $a_t$ at time step $t$ are no longer a sufficient statistic of history to predict the next observation and reward, $o_{t + 1}$ and $r_t$, or $Pr(o_{t + 1}, r_t \mid o_t, a_t) \ne Pr(o_{t + 1}, r_t \mid o_t, a_t, \dots, o_0, a_0)$. 
Under partial observability, an agent must use its history $h_t := (o_t, a_t, \dots, o_0, a_0) \in \mathcal{H}$ to learn a history-conditioned policy $\pi_{\Omega}: \mathcal{H} \rightarrow \Delta \mathcal{A}$ to maximize returns.

An agent can mitigate partial observability by learning memory functions $\mu: \mathcal{H} \rightarrow \mathbb{R}^n$. Memory functions condense past sequences of actions and observations into a memory state $\bm{m}_t = \mu(h_t)$. Since $h_t$ is variable in size, it is often more efficient and convenient to use recurrent memory functions $\bm{m}_t = \mu(o_t, a_t, \bm{m}_{t - 1})$. Ideally, a memory function learns to retain information that it needs in future decision making.
While traditional approaches have relied on discrete state machines to reason about states~\citep{chrisman1992fsm,peshkin1999memory},
most modern approaches leverage parameterized deep neural networks~\citep{goodfellow2016deeplearning} to learn memory functions.
One popular class of neural network memory functions are recurrent neural networks (RNNs)~\citep{amari1972rnn,mozer1995bptt}, powerful function approximators that can be optimized with truncated backpropagation through time~\citep{jaeger2002tbptt}.
Another state-of-the-art class of memory functions are transformers~\citep{vaswani2017attention}, which is not recurrent, and looks at a fixed context-length window of previous inputs in order to learn memory.
For reinforcement learning in partial observability, one can use standard gradient-based reinforcement learning algorithms to learn a neural network memory function capable of summarizing history to mitigate partial observability. 
The algorithm we use throughout this work for optimization is the popular proximal policy optimization algorithm (PPO)~\citep{schulman2017ppo}. We use this algorithm due to its strong performance in select partially observable environments with RNNs~\citep{Ni2021RecurrentMR} and transformers~\citep{ni2023when}. We also test on the $\lambda$-discrepancy algorithm~\citep{allen2024mitigating}, an extension to the recurrent PPO algorithm specifically made for mitigating partial observability.

There have been many forms of benchmark tasks for partial observability. Partially observable tasks were formulated to solve the POMDP planning problem~\citep{zhang2012covering}, the most well-known instance being the Tiger problem~\citep{Kaelbling98}. In most cases, the scale of these problems are too small and are easily approximated with modern neural networks~\citep{allen2024mitigating}. The few exceptions to this rule are benchmarks from POMDP planning algorithms designed to scale up to large state spaces~\citep{silver2010pomcp}, which we include in our study. Modern deep reinforcement learning algorithms have been tested on a number of difficult and large domains, including single-frame Atari~\citep{hausknecht15drqn}, masked~\citep{han2020variational} and visual~\citep{todorov2012mujoco,ortiz2024dmcvb} continuous control, and multiagent systems~\citep{flair2023jaxmarl,bettini2024benchmarl,LanctotEtAl2019OpenSpiel}. While there have been benchmarks specifically designed for partial observability~\citep{rajan2021mdp, morad2023popgym, osband2020bsuite}, these benchmarks tend to have a narrow range of partially observable tasks.

\section{Confounding Factors in Assessing Partial Observability Mitigation}
\label{sec:confounding_variables}

The objective of any benchmark is to give researchers a reasonable signal for progress on a class of problems. 
If the goal of an algorithm is to effectively mitigate partial observability, then progress measured in a benchmark should be from an agent effectively mitigating partial observability, as opposed to other factors.
While this may seem obvious, isolating performance increases is a challenging task in practice, considering how many factors affect deep reinforcement learning performance~\citep{henderson2018matters}. We begin by investigating some potential confounding factors in partially observable reinforcement learning.

\begin{figure*}[t]
    \centering
    \includegraphics[width=\linewidth]{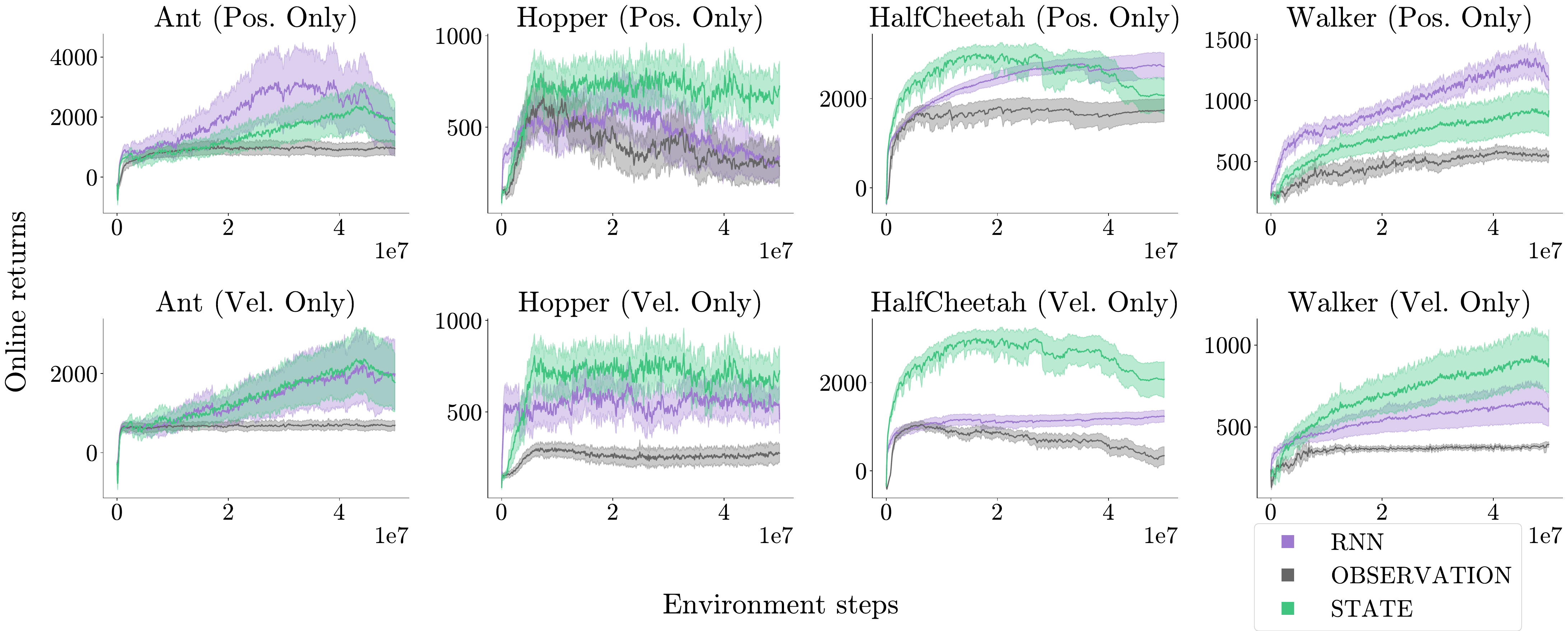}
    \caption{Masked continuous control online undiscounted returns for observations only (gray), full state (green), and an RNN agent (purple) over 30 seeds. Function approximation types play a large role in performance. Full experiment details are presented in Appendix~\ref{appx:masked_mujoco}.}
    \label{fig:masked_mujoco_results}
\end{figure*}

There are confounding factors in existing partially observable benchmarks that obfuscate the effects of partial observability.
In the Atari benchmark~\citep{bellemare13arcade}, a single frame is partially observable, whereas four stacked consecutive frames is usually assumed to be fully observable~\citep{mnih15dqn}. 
We would expect an agent imbued with state information to outperform an agent that receives only single frames and must do the extra work of resolving partial observability. In reality, results are much more complicated~\citep{hausknecht15drqn} and different algorithms make gains in different environments. 
In masked continuous control~\citep{han2020variational} one might expect an agent with full state features to perform better than one where certain features are masked out.
In Figure~\ref{fig:masked_mujoco_results} we show that more often than not, the opposite is true; RNNs under partial observability outperform memoryless agents with fully observable features, as with position-only Ant and Walker. 
It seems for most of these tasks, agents struggle with other factors besides a lack of information in the state representation.

Other confounding factors such as the choices of hyperparameters or function approximators
often impact performance in partial observability benchmarks. An important question to consider is: how much of the improvement is from mitigating partial observability and how much is from other factors?
Next, we study the effects of a few important general factors on performance for memory-learning tasks.

\subsection{Number of Parallel Environments}
\label{sec:ablation_nenv}
\begin{figure}[t]
    \centering
    % \vspace{-20px}
    \begin{subfigure}[b]{0.32\linewidth}
        \begin{minipage}{\linewidth}
            \centering
            \includegraphics[width=\linewidth]{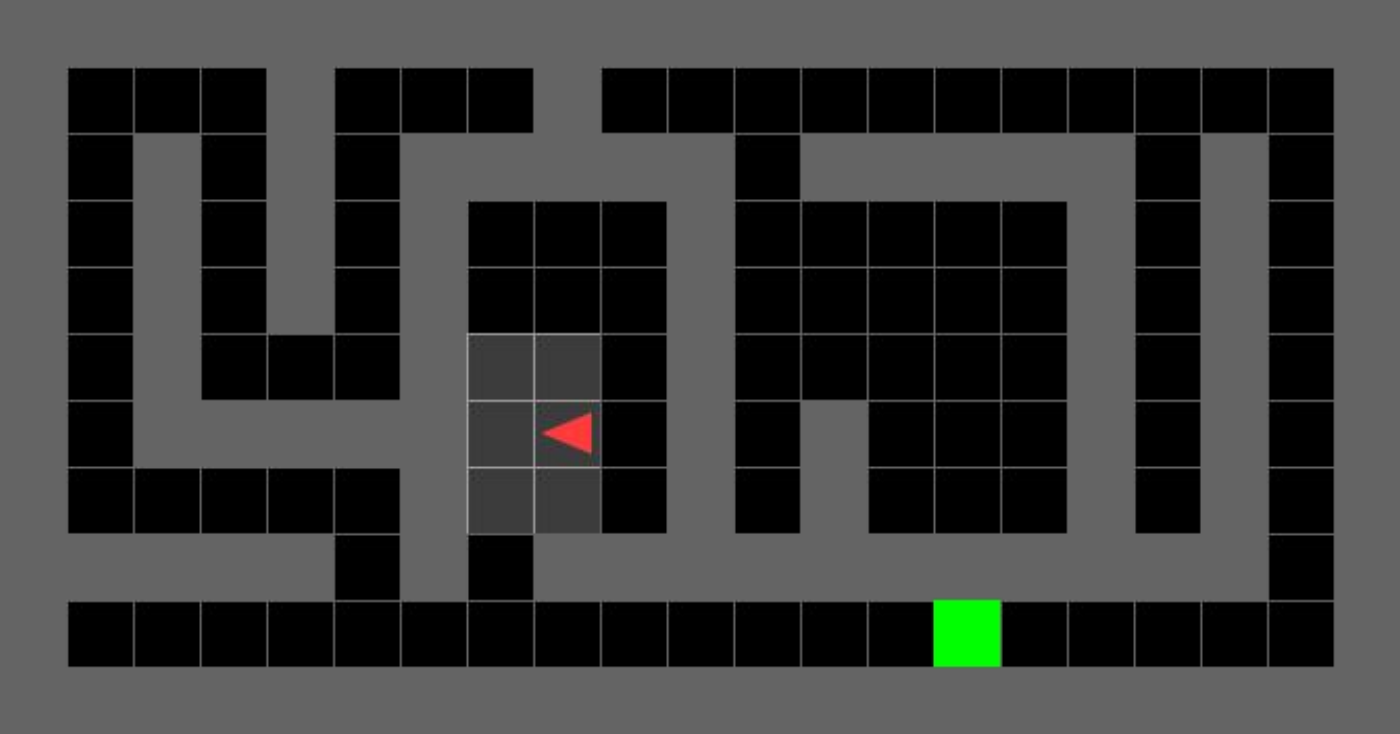}
        \end{minipage}
    \end{subfigure}
    \hfill
    \begin{subfigure}[b]{0.64\linewidth}
        % \vspace{22px}
        \begin{minipage}{\linewidth}
            \centering
            \includegraphics[width=\linewidth]{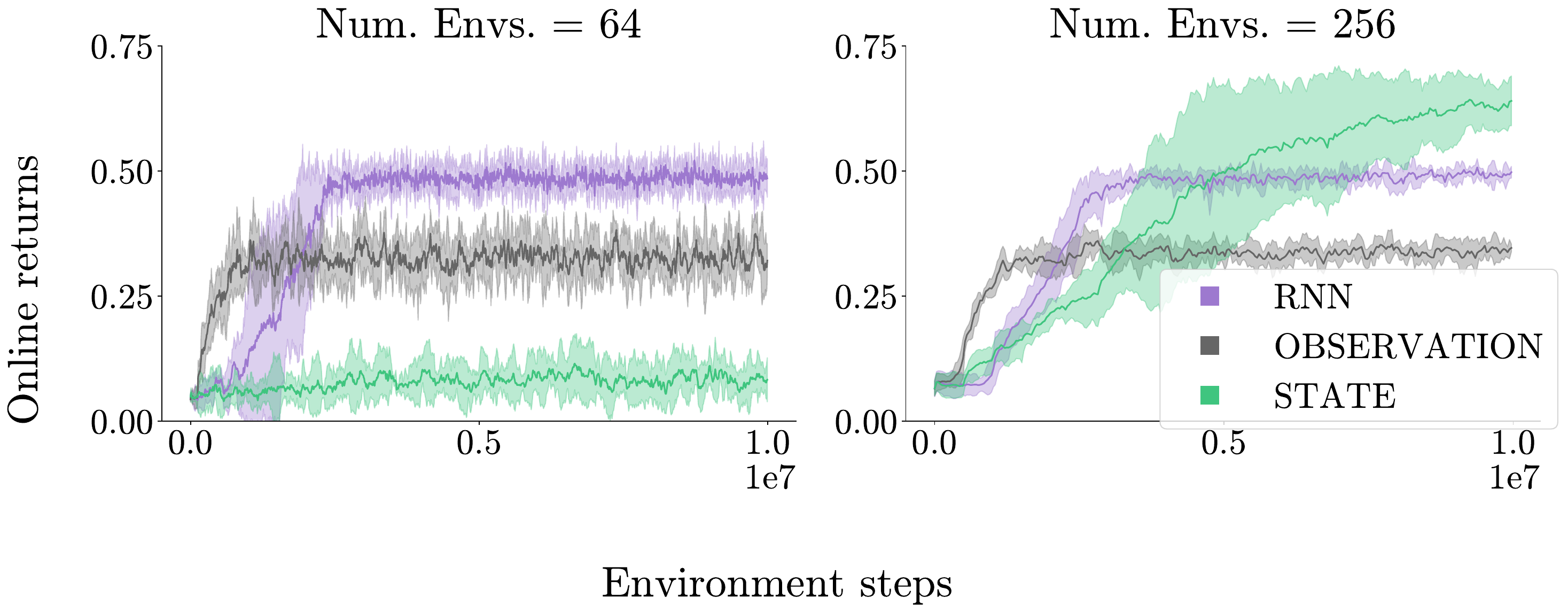}
        \end{minipage}
    \end{subfigure}
    \caption{(Left) Image of the DMLab Minigrid maze environment for \texttt{maze\_id = 01}. (Middle, right) Online discounted returns in this environment comparing performance of using 64 vs 256 parallel environments. Experiments were conducted over 5 seeds.}
    \label{fig:ablation_parallel_envs}
\end{figure}

Modifying the number of parallel copies of environments can drastically change the performance of a given featurization and algorithm.
Reinforcement learning algorithms will use parallel copies of an environment to make uncorrelated minibatches of experience for more stable gradient updates.
Figure~\ref{fig:ablation_parallel_envs} shows an ablation study on the number of parallel environments in the DeepMind Lab Minigrid domain introduced in Section~\ref{sec:pobax_envs}. 
Note that the total number of environment steps used for training remains the same. The difference is in the size of the minibatch for each gradient update. As the number of parallel environments increases, the size of each minibatch increases, but the number of total updates decreases.
We generally see improved performance with an increase in the number of parallel environments. Full details of this ablation study are in Appendix~\ref{appx:ablations}.
The trade-off for increasing the number of parallel environments is increased memory usage, making experiments less scalable with more parallel environments. To ameliorate this variance, the benchmark we introduce includes recommendations for the number of parallel environments required for each task such that our baseline and skyline agents both learn.

\subsection{Network Width}
\label{sec:ablation_hsize}

\begin{figure}[t]
    \centering
    \includegraphics[width=0.95\linewidth]{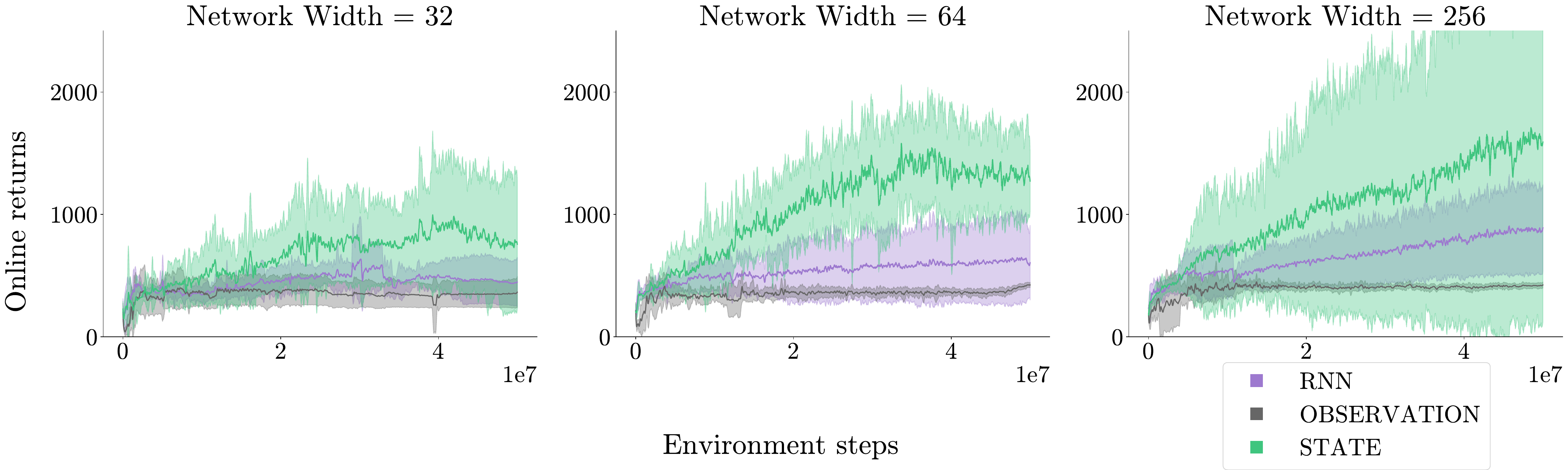}
    \caption{Online undiscounted returns comparing network hidden sizes 32, 64 and 256 (left to right) on velocity-only Walker.}
    \label{fig:ablation_net_width}
\end{figure}

Network width is another general hyperparameter for deep reinforcement learning agents with a sizable but diminishing effect as width increases. The network width is the number of neurons in a neural network's hidden layers, also called its hidden size. In Figure~\ref{fig:ablation_net_width} investigate the effect of network width for the velocity-only Walker environment from the masked continuous control benchmark. As network width increases, we see consistent but diminishing improvements in performance. 
The trade off with increased network width is again a large computational and memory overhead, requiring more resources per experiment. Our benchmark also includes default recommended network widths for each environment. All details of this ablation study are shown in Appendix~\ref{appx:ablations}.

We advocate for choosing general hyperparameter settings for each environment and fixing these settings across all algorithms to ensure a fair comparison between algorithms.
Ideally, these settings should also be swept for each algorithm; but with computational resource constraints, sweeping many settings is untenable.
As an alternative, our proposed benchmark provides recommended settings for general hyperparameters, including the two studied in this section. This is not to say practitioners should stop sweeping more algorithm-dependent hyperparameters like learning rate. Instead we advocate for a reasonable middle ground for computational feasibility and experimental rigor.
Beyond these two hyperparameter settings, many other factors can affect deep reinforcement learning performance. 
Input featurization and neural network normalization are just a few factors important for performance that we do not investigate in this work.
Fixing these confounding factors, we now consider properties that make for a good partially observable benchmark.

\section{Memory Improvability}
\label{sec:mem_improve}

Controlling for confounding factors is not enough to isolate performance gains from mitigating partial observability.
We argue that the most important characteristic of an environment is its memory improvability: an indication that performance gains are likely from mitigating partial observability. An environment is memory-improvable if there exists a gap between the performance of agents with less or more state information. If this gap exists, assuming most other factors are equal (e.g. learning algorithm, network size), 
then gains from a memory-learning algorithm will likely come from mitigating partial observability.

\begin{figure*}[t]
    \centering
    \includegraphics[width=0.7\linewidth]{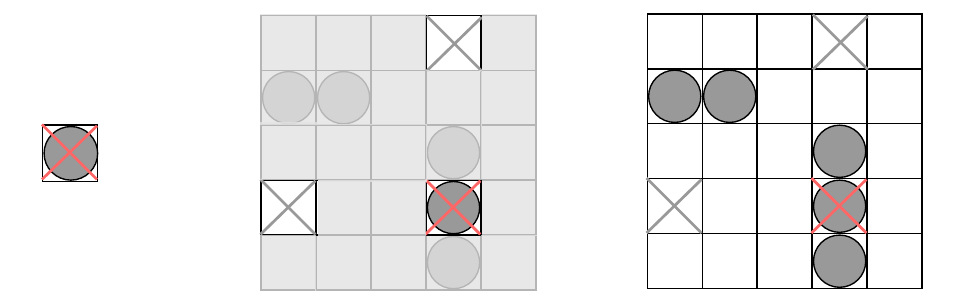}
    \caption{Different levels of observability in $5 \times 5$ \textit{Battleship}. (Left) Observations in this version of Battleship are whether or not the previous action hit. (Middle) ``Perfect memory'' observability, where observations include all previous position hit and missed. Grayed out grids are unobservable. (Right) Full observability, where ship positions are also included in observations.}
    \label{fig:observability_examples}
\end{figure*}

Environments should therefore admit multiple state representations that contain differing amounts of state information such that merely adapting the agent to the new input space is sufficient to achieve a performance improvement with minimal algorithmic changes.

Consider the forms of observability in a version of the game \textit{Battleship}~\citep{silver2010pomcp} in Figure~\ref{fig:observability_examples}. 
In this game, players must select coordinates on a grid to fire at in order to sink ships. We show three examples of observability here: first is the least observable version, where observations only include whether or not the last shot hit. This poses a particularly hard challenge, since in addition to learning the dynamics of Battleship, the agent must also remember previous shot locations. The second agent has ``perfect memory'' where the observation is Markov (since no additional information can be gleaned from previous observations) and all previous hits and misses are tabulated in a grid. Lastly, we have the full state observation that also includes ship positions. 
An agent that learns memory should be able to attain performance matching an agent with perfect memory, whereas optimal performance with full observability is an upper-bound for performance, oftentimes unachievable.
Performance with base memoryless observations gives a floor to the performance of an agent, whereas performance with either perfect memory or full state observations gives a ceiling. 
If a gap exists between the performance of these agents, then an environment is memory improvable. 
Conversely, it is also possible to create a memory improvability gap by further reducing the amount of information in already-partially-observable state features; for example, features in Battleship that only reveal hits but not misses in Battleship.

With other factors held constant, performance gains by a partial-observability-mitigating algorithm in a memory-improvable environment are more likely due to mitigating partial observability.
From Section~\ref{sec:confounding_variables} we know that without memory improvability, performance gains on a partially observable domain could be due to other confounding factors. We discuss some differences between our results and those seen in other works from uncontrolled confounding factors in Appendix~\ref{appx:differences_benchmarks}. When the biggest difference between agents is the information in the input features, the gains above the agent with less information are more likely from an agent better mitigating partial observability.

Now that we have described how we intend to evaluate agents with our benchmark, we can assess which environments would make for a good evaluation for mitigating partial observability.

\section{Categorizing Partial Observability}
\label{sec:categorizing_po}

To choose representative environments for benchmarking partial observability, we first must define categories of interest that partially observable environments fall into. 
In the following list, we focus on the different forms that partial observability can take, as opposed to categorization with solution methods in mind.
% We also emphasize that this is in no way an exhaustive list of 
We define eight categories popular in partial observability and example problems for each. Note that environments may fall into multiple categories of partial observability. We emphasize that this is not an exhaustive list of the archetypes of partial observability, but merely popular forms seen throughout reinforcement learning literature.

\paragraph{Noisy state features} 
State features with additive noise. The most popular option for additive noise is to add Gaussian noise to continuous state features: $\phi(\bm{x}(s)) := \bm{x}(s) + \delta$, where $\delta$ is sampled noise from a multivariate Gaussian with zero mean. Modeling partial observability as additive Gaussian noise is a popular technique in robotics~\citep{thrun2005robots}. An example of this is noisy Cartpole and Pendulum environments~\citep{morad2023popgym}, where baseline observation-only agents already perform well. Additive state features may not provide the best signal for algorithmic progress in partial observability.

\paragraph{Visual occlusion} 
% Example: visual mujoco
A portion of the environment's visibility is occluded by other parts of the environment or distance. Visual occlusion is one of the most popular sources of partial observability in both robotics and reinforcement learning, with visual locomotion~\citep{todorov2012mujoco} and occluded maze navigation~\citep{beattie2016dmlab, boisvert2023minigrid} as popular and challenging existing benchmarks. 

\paragraph{Object uncertainty \& tracking}
% example: rocksample
The state of objects in the environment are unknown, requiring an agent to reason about each object and potentially track it. The classic POMDP benchmark RockSample~\citep{smith2004rocksample} is an apt example, since an agent must test and remember the parity of each rock. Games such as Crafter~\citep{hafner2021crafter} contain objects and enemies that may leave the screen which an agent should track or act to observe.

\paragraph{Spatial uncertainty}
% Example: slam in general, partially observable pacman
Environments where the agent is required to localize and potentially map its environment. This form of partial observability is a classic task in robotics~\citep{thrun2005robots}. In reinforcement learning, the aforementioned maze navigation~\citep{beattie2016dmlab} and first-person grid world environments~\citep{boisvert2023minigrid,pignatelli2024navix} are popular examples.

\paragraph{Moment features}
% example: masked mujoco. WHat's interesting is positional masked mujoco.
Environments where state representation is characterized by moments. In continuous control domains~\citep{todorov2012mujoco}, position and velocity (first and second moments) of the agent's joints characterize the full state of the system. Environments can be made partially observable by obscuring position or velocity information~\citep{han2020variational}.

\paragraph{Unknown opposition}
% Example: prisoner's dilemma/many multiagent systems. We don't include these 
In multiagent systems, an agent is unaware of the opponent's policy, making the world partially observable. Adding more agents, each with their own policy, exponentially increases the size of the system. Multiagent reinforcement learning is a large field of study with many existing benchmarks~\citep{flair2023jaxmarl,bettini2024benchmarl,LanctotEtAl2019OpenSpiel}. Due to the scope of this category, we leave this form of partial observability to these benchmarks.

\paragraph{Episode nonstationarity}
% Example: dmlab mazes with random starts/ends
Tasks where aspects of the environment change over episodes. Maze environments from DeepMind Lab~\citep{beattie2016dmlab} are a classic example of this, where the start and goal positions are randomized at every step for each maze configuration. ProcGen~\citep{cobbe2019procgen} is an extreme example of this, the environment is partially observable and each episode also instantiates in a randomly generated level of each game.

\paragraph{Needle in a haystack}

These difficult environments test an agent's ability to memorize a random sequence of events, oftentimes unrelated to one another. An example of this is the diagnostic Autoencode task~\citep{morad2023popgym}, where an agent must repeat back a shuffled deck of 52 cards backwards. In this setting, the only sequence of observations that holds any information about rewards is the sequence of cards shown to the agent---there is no accumulation of information, only a single sequence of actions among exponentially many possibilities of sequences that will result in a reward. We leave out environments of this form because they are diagnostic and purely meant to test memory length, as opposed to partial observability of interest.

Together with memory improvability in Section~\ref{sec:mem_improve}, we are now ready to establish a benchmark for mitigating partial observability.

\section{POBAX: A Fast, Memory-Improvable Benchmark for Reinforcement Learning Under Partial Observability}

\textbf{Partially Observable Benchmarks in JAX (POBAX)} is a new suite of reinforcement learning environments for benchmarking partial observability.
It includes partially observable environments with hard-to-learn memory functions. These environments cover the categories of partial observability of interest in Section~\ref{sec:categorizing_po}, and are all memory improvable with the provided recommended hyperparameter settings.
POBAX is also written entirely in JAX~\citep{jax2018github} which allows for fast GPU-scalable experimentation. Timing experiments comparing GPU-accelerated POBAX environments to regular Gymansium~\citep{towers2024gymnasium} environments are presented in Appendix~\ref{appx:gpu_scalability}.

\subsection{Environments}
\label{sec:pobax_envs}

\begin{figure}[t]
    \centering
    % \vspace{-20px}
    \begin{subfigure}[b]{0.28\linewidth}
        % \vspace{22px}
        \begin{minipage}{\linewidth}
            \centering
            \begin{subfigure}[b]{\linewidth}
                \begin{minipage}{\linewidth}
                    \centering
                    \includegraphics[width=\linewidth]{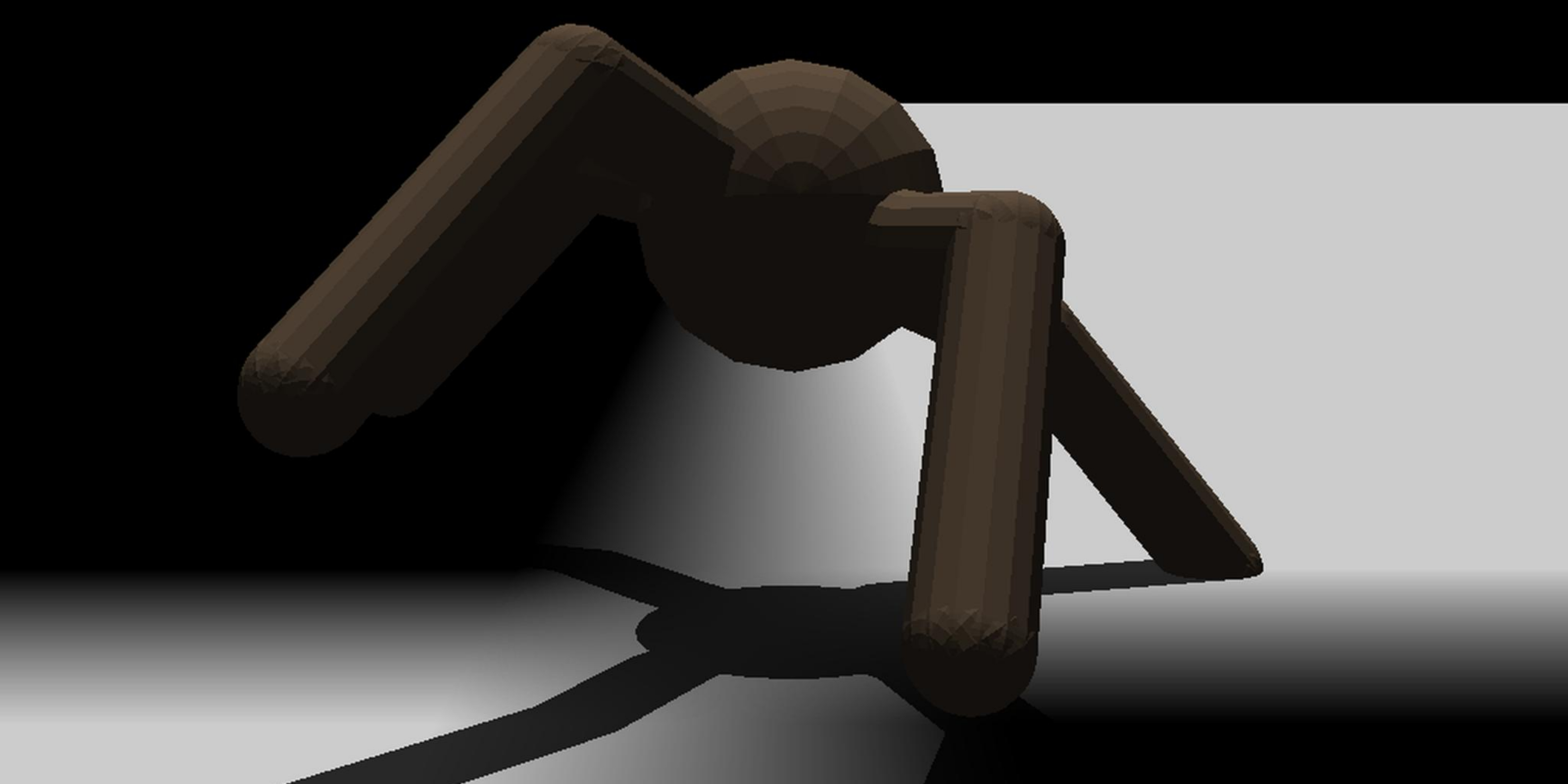}
                \end{minipage}
            \end{subfigure}
            \begin{subfigure}[b]{\linewidth}
                \begin{minipage}{\linewidth}
                    \centering
                    \includegraphics[width=\linewidth]{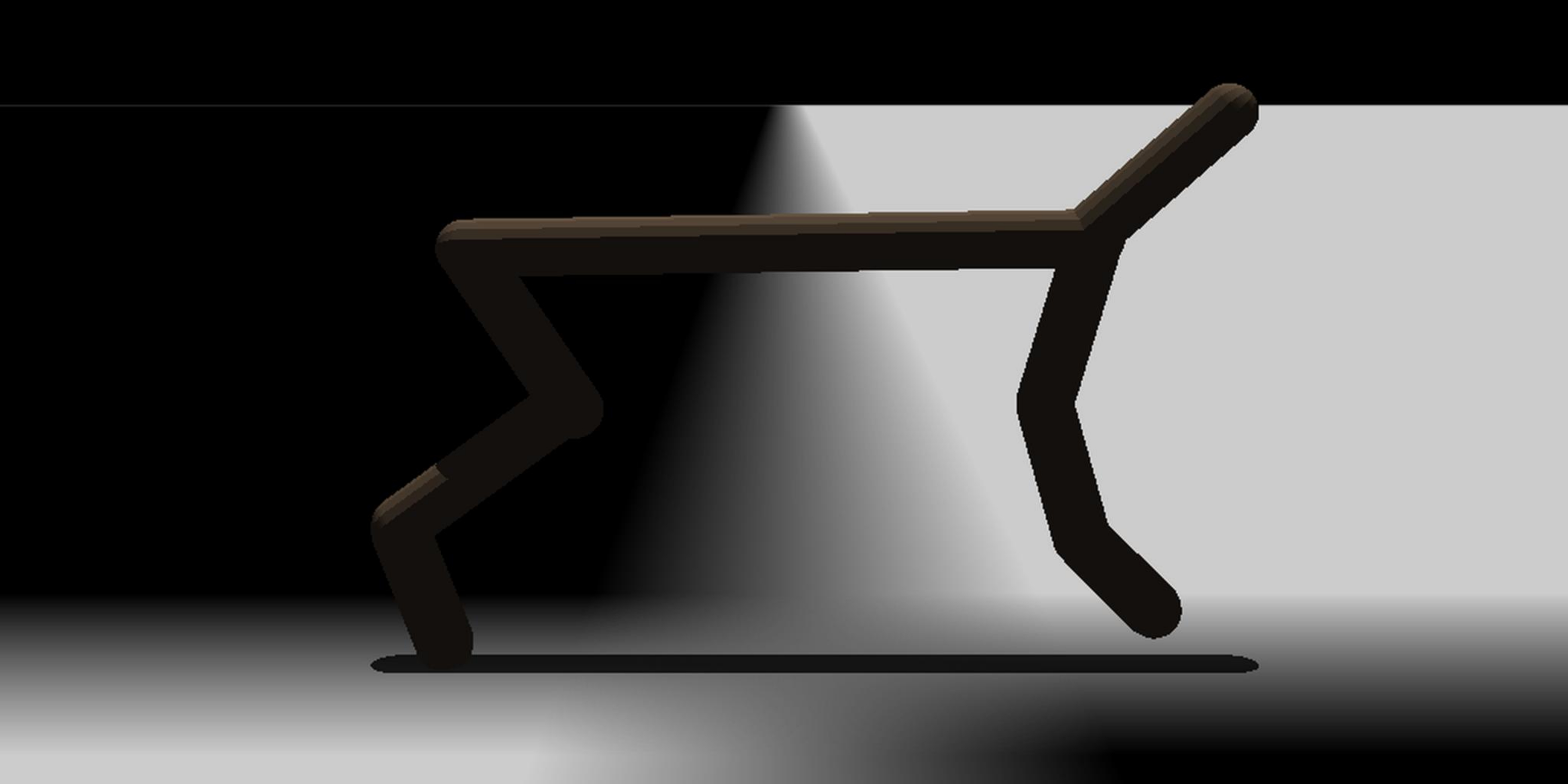}
                \end{minipage}
            \end{subfigure}
        \end{minipage}
        \caption{Visual Mujoco}
        \label{fig:env_images_mujoco}
    \end{subfigure}
    \hfill
    \begin{subfigure}[b]{0.57\linewidth}
        \begin{minipage}{\linewidth}
            \centering
            \includegraphics[width=\linewidth]{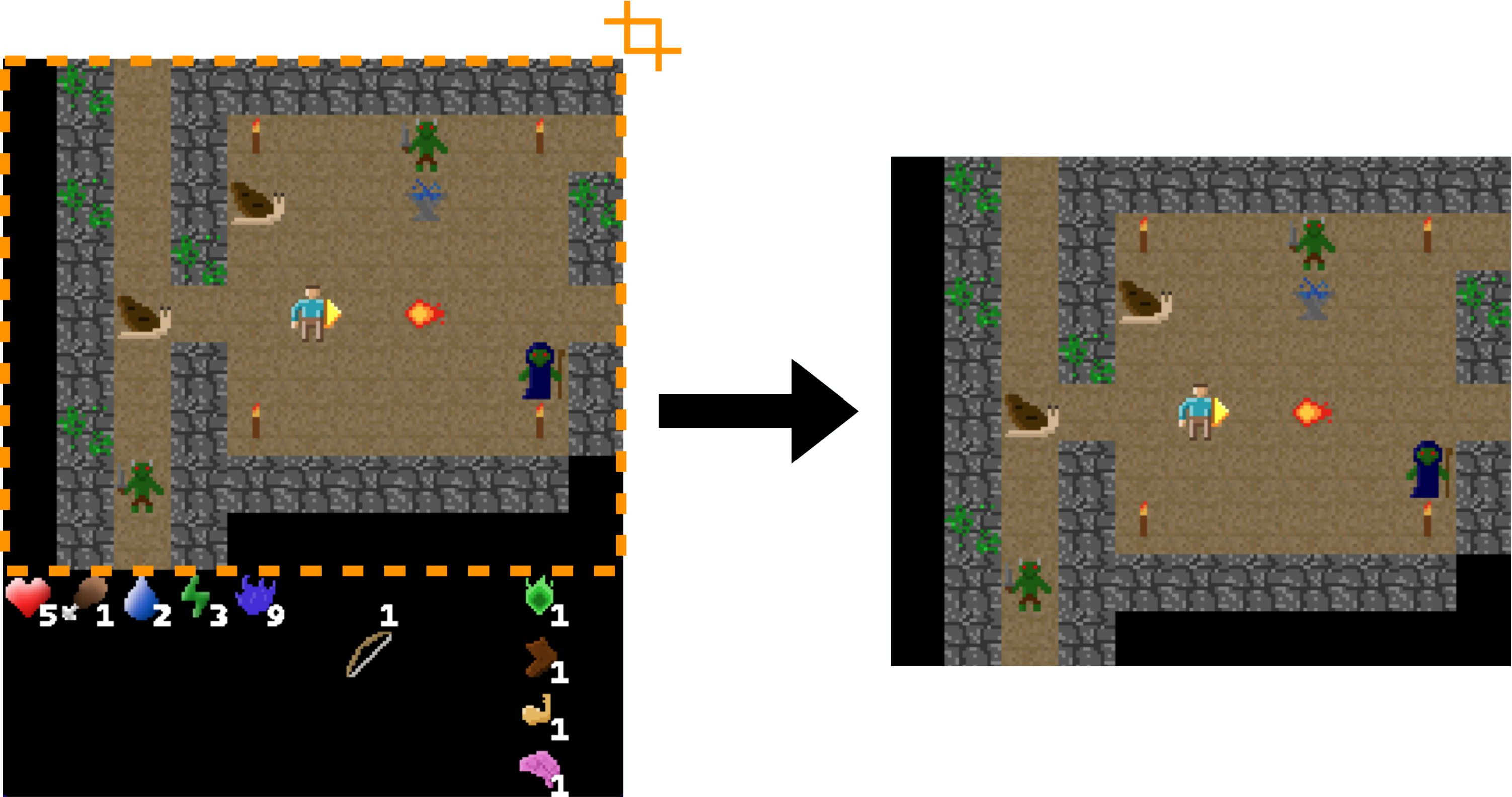}
        \end{minipage}
        \caption{No-inventory Crafter}
        \label{fig:env_images_crafter}
    \end{subfigure}
    \caption{Pixel-based environments in POBAX. (Left) Ant and HalfCheetah in visual continuous control. Images are rendered with full JAX support in the Madrona MJX rendering engine~\citep{Shacklett2024madrona}, with the dark coloration due artifacts of the new framework. (Right) Observations in no-inventory Crafter have the agent's inventory cropped out, requiring the agent to remember its items and stats. }
    \label{fig:env_images}
\end{figure}

We briefly summarize each environment before testing them on a set of popular reinforcement learning algorithms made for mitigating partial observability. Environment identification strings (for the \texttt{get\_env} function) are given after their names. Full details of all environments are in Appendix~\ref{appx:env_and_hyper_details}.

\paragraph{T-Maze}(\texttt{tmaze\_\{$n_{\text{length}}$\}}) A small diagnostic benchmark for partial observability and memory length~\citep{bakker2001reinforcement}. 
At the beginning of an episode, the agent is told whether the reward at the end of a hallway is up or down, and the agent must remember this by the time it gets to the T-junction.
We recommend using this environment as a sanity check for memory learning algorithms, since the optimal policy's return will always be $4 \times \gamma^{n_{\text{length}} + 1}$, where $n_{\text{length}}$ is the length of the hallway. Category: \textit{object uncertainty \& tracking}

\paragraph{RockSample}(\texttt{rocksample\_11\_11} and \texttt{rocksample\_15\_15}) A classic medium-sized problem in POMDP literature~\citep{smith2004rocksample}. 
In \textit{RockSample(11, 11)} and \textit{RockSample(15, 15)}, the agent needs to sample good rocks throughout its environment and exit. Partial observability comes from the need to test each rock with its distance-dependent stochastic sensor. This environment is extendable to the general RockSample($n_{\text{grid}}$, $k$) problem, where $n_{\text{grid}}$ is the size of the $n_{\text{grid}} \times n_{\text{grid}}$ grid, and $k$ is the number of randomly dispersed rocks in the environment. Category: \textit{object uncertainty}

\paragraph{Battleship}(\texttt{battleship\_10}) Another medium-sized problem based on the board game, also from POMDP planning literature~\citep{silver2010pomcp}. An agent must hit all 4 ships in a $10 \times 10$ grid, and sees only \texttt{HIT} or \texttt{MISS} at every step. This environment is extendable to any $n_{\text{grid}} \times n_{\text{grid}}$ map, with any number of ships of any sizes. Categories: \textit{spatial uncertainty} and \textit{episode nonstationarity}

% \paragraph{P.O. PacMan}(\texttt{pocman}) Partially observable version of PacMan. A medium-sized problem used to test large-scale POMDP planning~\citep{silver2010pomcp}. In this version of PacMan, the agent can only see its surrounding walls and pellets, as well as if a ghost is in its field-of-view. This environment was built on top of the PacMan game from the Jumanji framework~\citep{bonnet2024jumanji}. Categories: \textit{spatial uncertainty} and \textit{object uncertainty \& tracking}

\paragraph{Masked Mujoco}(\texttt{Walker-V-v0} and \texttt{HalfCheetah-V-v0}) Medium-sized continuous control environments (Walker and HalfCheetah) with only velocity features~\citep{han2020variational}. In this setting, an agent is required to integrate over its history of velocities to mitigate partial observability. From the experiments in Figure~\ref{fig:masked_mujoco_results}, both \texttt{Walker-V-v0} and \texttt{HalfCheetah-V-v0} are memory improvable and we include them in this benchmark. These environments were made on top of the Brax framework~\citep{brax2021github}. Category: \textit{moment features}

\paragraph{DeepMind Lab MiniGrid mazes} (\texttt{Navix-DMLab-Maze-\{maze\_id\}-v0}, $\texttt{maze\_id} \in \{\texttt{01}, \texttt{02}, \texttt{03}\}$) Medium-to-large tasks that are 2D versions of the DeepMind Lab~\citep{beattie2016dmlab} mazes implemented in MiniGrid~\citep{boisvert2023minigrid, pignatelli2024navix}, as seen in Figure~\ref{fig:ablation_parallel_envs}. The agent is randomly initialized to a start position and has to navigate to a randomly sampled goal position. Observations are agent-centric views of the $3 \times 2$ area in front of itself, requiring an agent to localize in its environment and find where the goal is. This environment was built on top of the NAVIX framework~\citep{pignatelli2024navix}. Categories: \textit{spatial uncertainty} and \textit{episode nonstationarity}.

\paragraph{Visual Mujoco}
(\texttt{ant\_pixels} and \texttt{halfcheetah\_pixels}) Large-scale continuous control with single-frame observations~\citep{todorov2012mujoco}. An agent is required to gauge its proprioceptive state through frame-by-frame pixel images, as shown in Figure~\ref{fig:env_images_mujoco}. Using pixel images not only obfuscates the velocity of each joint, but also includes visual occlusion of the other aspects of the state. These environments were built on top of the Brax framework~\citep{brax2021github}. Categories: \textit{visual occlusion} and \textit{moment features}.

\paragraph{No-inventory Crafter}(\texttt{craftax\_pixels})
Large-scale pixel-based alternative version of the Crafter benchmark~\citep{hafner2021crafter}. In regular Crafter, the agent is already partially observable. To make a memory improvability gap, we make the original state features more partially observable by obscuring the agent's inventory as shown in Figure~\ref{fig:env_images_crafter}. This version, called no-inventory Crafter, is memory improvable because there is a performance gap between the original Crafter observations and the no-inventory observations. This environment was built on top of the Craftax framework~\citep{matthews2024craftax}. Categories: \textit{visual occlusion}, \textit{spatial uncertainty}, and \textit{object uncertainty \& tracking}.

\begin{figure*}[t]
    \centering
    \includegraphics[width=0.9\linewidth]{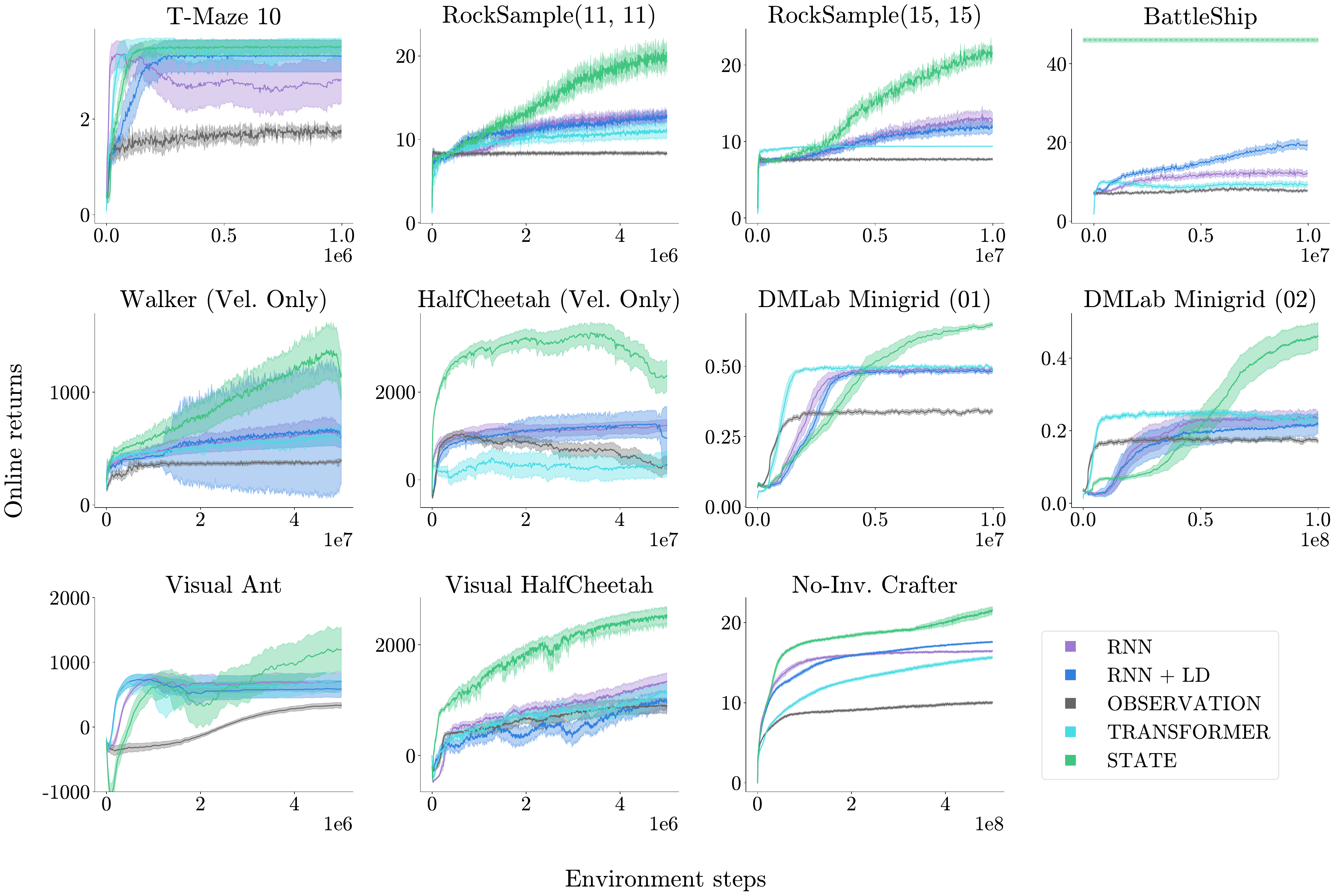}
    \caption{Performance across all POBAX domains. Experiments are run over 30 seeds, with shaded regions denoting a $95\%$ confidence interval.}
    \label{fig:all_envs_results}
\end{figure*}

\subsection{Results}
We test the above environments on three popular reinforcement learning algorithms designed for mitigating partial observability:
\begin{enumerate}
    \item Recurrent PPO~\citep{schulman2017ppo},
    \item $\lambda$-discrepancy~\citep{allen2024mitigating} with recurrent PPO,
    \item Transformer-XL~\citep{parisotto2020transformerxl} with PPO.
\end{enumerate}

General hyperparameters for each environment were kept fixed, while algorithm-specific hyperparameters were swept. Both recommended environment hyperparameters and swept-and-selected algorithm hyperparameters are detailed in Appendix~\ref{appx:hyperparam_details}.

To show the utility of our library, we evaluate all three memory-based reinforcement learning algorithms on the POBAX benchmark environments listed in Section~\ref{sec:pobax_envs}. Results are shown in Figure~\ref{fig:all_envs_results}.
The gap between observations-only agents (gray) and the additional state information agents (green) imply that the environments are all memory improvable. 
All three memory-learning algorithms manage to improve upon the performance of the observations-only agent, and underperform the agent with more state information, implying that performance gains are most likely from mitigating partial observability.
Results show mean and $95\%$ confidence interval over 30 seeds.

Ceilings for performance do not have to be the ``perfect memory'' or fully observable featurizations.
In no-inventory Crafter, the ``full state'' agent is in fact a transformer agent trained on regular Crafter with the inventory included in the observations. 
The full state agent in BattleShip is the mean performance of an optimal belief policy~\citep{berry2011battleship} calculated programmatically. Both of these ceilings represent the mean performance that an algorithm with its original observation feature set should be able to achieve if it can mitigate partial observability effectively.

Finally, the third DMLab Minigrid maze (\texttt{maze\_id = 03}) was not included in this benchmark due to its difficulty. In addition to requiring complex localization of the environment, these maze environments also pose a hard exploration and sparse reward task for all three algorithms. 
For \texttt{maze\_id = 01, 02}, agents were trained on 256 and 512 parallel environments respectively in order for agents to learn effectively. This large number of parallel environments already pose a significant computational overhead, leaving the third task as a difficult, unsolved challenge.

\section{Conclusion}

Benchmarking an algorithm's ability to mitigate partial observability is challenging due to the scope that partial observability covers and
the many confounding factors of deep reinforcement learning.
We introduce POBAX: Partially Observable Benchmarks for reinforcement learning in JAX. This open-source benchmark is built around two key properties: coverage over many forms of partial observability and memory improvability. 
An environment is memory improvable if performance gains are from an algorithm's ability to mitigate partial observability as opposed to other factors.
To achieve memory improvability in our benchmark, we investigate the affects of different confounding factors on performance to give a recommended set of hyperparameters for each environment. 
We then introduce categories of partial observability of interest and select representative environments for our benchmark. 
Experimental results show that the POBAX benchmark environments are memory improvable, and evaluation of three popular algorithms demonstrate the utility of the benchmark as a signal for research on mitigating partial observability in reinforcement learning.

%%%%%%%%%%%%%%%%%%%%%%%%%%%%%%%%%%%%%%%%%%%%%%%%%%%%%%%%%%%%%%%%
%% Appendices
%%%%%%%%%%%%%%%%%%%%%%%%%%%%%%%%%%%%%%%%%%%%%%%%%%%%%%%%%%%%%%%%
% \appendix

% \section{The first appendix}
% \label{sec:appendix1}
% This is an example of an appendix. 

% \noindent \textbf{Note:} Appendices appear before the references and are viewed as part of the ``main text'' and are subject to the 8--12 page limit, are peer reviewed, and can contain content central to the claims of the paper. 

\newpage

\subsubsection*{Acknowledgments}
\label{sec:ack}
We would like to thank Ron Parr and our colleagues at Brown University for their valuable feedback during the development of the benchmark. We would also like to thank the Reinforcement Learning Conference reviewers for their numerous suggestions that has improved the work. This work was generously supported by the Office of Naval Research (ONR) ONR grant \#N00014-22-1-2592.
% Use unnumbered third level headings for the acknowledgments. All acknowledgments, including those to funding agencies, go at the end of the paper. Only add this information once your submission is accepted and deanonymized. The acknowledgments do not count towards the 8--12 page limit.

%%%%%%%%%%%%%%%%%%%%%%%%%%%%%%%%%%%%%%%%%%%%%%%%%%%%%%%%%%%%%%%%
%% NOTE: THIS MARKS THE END OF THE "MAIN TEXT"
%%%%%%%%%%%%%%%%%%%%%%%%%%%%%%%%%%%%%%%%%%%%%%%%%%%%%%%%%%%%%%%%

%%%%%%%%%%%%%%%%%%%%%%%%%%%%%%%%%%%%%%%%%%%%%%%%%%%%%%%%%%%%%%%%
%% Bibliography
%%%%%%%%%%%%%%%%%%%%%%%%%%%%%%%%%%%%%%%%%%%%%%%%%%%%%%%%%%%%%%%%
\bibliography{main}
\bibliographystyle{rlj}

%%%%%%%%%%%%%%%%%%%%%%%%%%%%%%%%%%%%%%%%%%%%%%%%%%%%%%%%%%%%%%%%
% AUTHOR: If your paper has no supplementary materials, you may 
%         comment out the line below, which creates the title for
%         the supplementary materials.
%%%%%%%%%%%%%%%%%%%%%%%%%%%%%%%%%%%%%%%%%%%%%%%%%%%%%%%%%%%%%%%%
\beginSupplementaryMaterials

\appendix
\section{Environment GPU Scalability Experiments}
\label{appx:gpu_scalability}

\begin{figure*}[h]
    \centering
    \includegraphics[width=\linewidth]{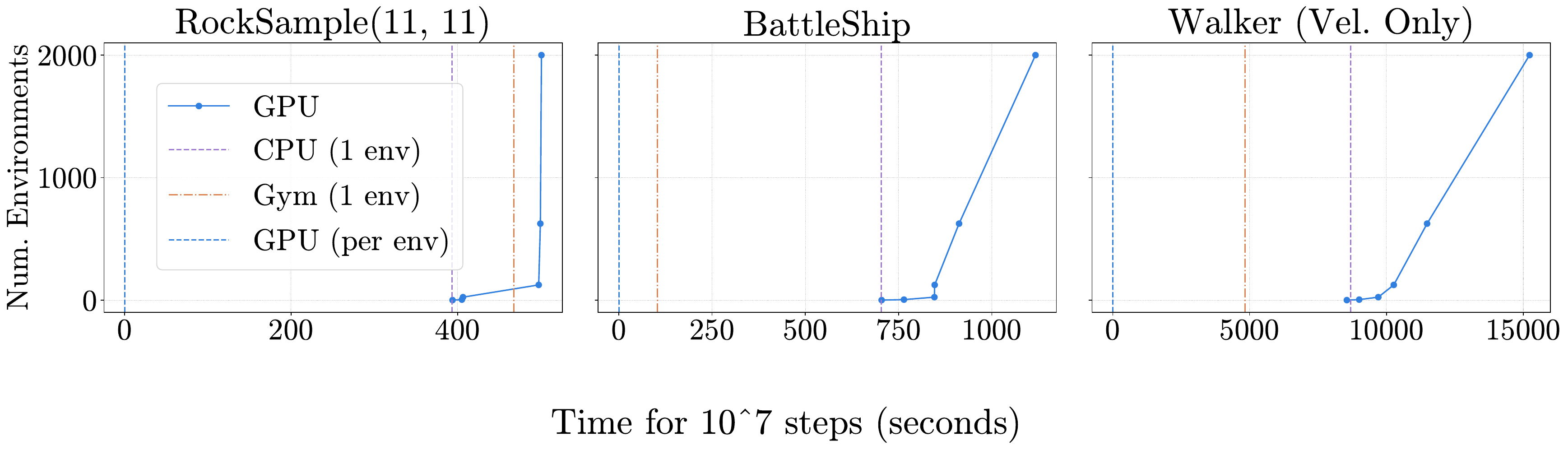}
    \caption{Wall clock speeds over number of parallel environments run for 10M steps. Dashed vertical lines represent the time it takes to run 10M steps in a single environment for the JAX environment on CPU (purple) and a single environment for an equivalent Gymnasium~\citep{towers2024gymnasium} implementation (orange). The dashed blue line represents the time it takes per environment for 10M steps when running 2000 environments on GPU.}
    \label{fig:env_timing}
\end{figure*}

We compare wall-clock speeds for our JAX-implemented environments versus pure Python implemented versions of the same environments in Figure~\ref{fig:env_timing}. Experiments were conducted on an NVIDIA 3090 GPU and AMD 5900X CPU.

The scaling curve for the GPU accelerated runs are desirable for large-scale experimentation. In this setting, each run may run with multiple parallel environments, with multiple seeds and multiple hyperparameter configurations, resulting in thousands of individual environments running for just a single environment. The GPU curve (blue solid line) scales particularly well with number of environments, as seen by the steep timing curve. While the Gymnasium environments (dashed vertical orange line) perform better in some cases than a single JAX environment (dashed vertical purple line for CPU), the per-environment time to run 10M steps when scaling on GPU over 2000 environments (dashed vertical blue line) is by far the fastest.

\section{Algorithmic and Hyperparameter Details}
We now detail all algorithmic and architectural details in our deep reinforcement learning experiments.

\label{appx:hyperparam_details}
\subsection{Algorithms}
% No label, since this can't be referenced meaningfully with \ref{}.
Our base PPO algorithm is an online learning method designed for training on vectorized environments. It is parallelized using the JAX library \citep{jax2018github} based on a batch experimentation library written in JAX \citep{lu2022discovered}. 

Our experiments consist of two steps. First, we perform a hyperparameter sweep over all environments, using a small number of seeds. Then, we select the best hyperparameters based on the highest area under the curve (AUC) score. After selection, we rerun the best hyperparameters using 30 seeds to generate our results. Note that specific hyperparameters being swept and the number of seeds may vary depending on the domain.

We evaluate four algorithms: Memoryless PPO, Recurrent PPO, $\lambda$ discrepancy with Recurrent PPO, and Transformer-XL. Memoryless PPO is the standard PPO algorithm without any form of internal memory. This means that it solely relies on the current observation to make decisions. In contrast, the other three algorithms are all memory learning algorithms which incorporate a mechanism to capture past experiences. We use observations-only PPO to show the memory improvability gap in our environments, we run this algorithm twice---once on partial observations and once on full observation to acquire the floor and ceiling of our plots. For Recurrent PPO, $\lambda$ discrepancy with Recurrent PPO, and Transformer-XL, we also concatenate our action into the observation, which provides additional context to enhance memory learning.

We implement our recurrent PPO model following the approach detailed in \citep{lu2022discovered} and we implement the $\lambda$-discrepancy algorithm following the implementation of \citep{allen2024mitigating}.

Transformer-XL is a memory-augmented algorithm that extends from the conventional architecture of transformers by incorporating segment-level recurrence. Our algorithm followed the implementation of \cite{hamon2024trxl}. One thing to notice is that traditional transformers use its attention mechanism on a fixed input sequence, during which it will lose temporal information and limit their ability to capture dependencies that span beyond the current window. Transformer-XL overcomes this by storing the hidden states from previous sequence, effectively extending the window of information to allow the agent to acquire information from earlier observations. 

\subsection{Network Architecture}
The general architecture of the network used in all our experiments consist of three parts. First, if the environments have visual inputs, we use either FullImageCNN or SmallImageCNN. Then, we get the feature representations by one of three modules—Memoryless, Recurrent Neural Network (RNN), or Transformer—depending on the algorithms. Finally, we called Actor Critic on the processed features for decision making. The detailed descriptions of the components are provided in the following paragraphs.

\paragraph{Actor Critic}All our models use an actor-critic architecture. Both actor and critic networks consist of two layers of standard multi-layer perceptron (MLP) with ReLU activations between layers. There is an additional Categorical or MultivariateNormalDiag functions applied at the end of actor network over actor logits depending on the action space of the environments. 

\begin{verbatim}
ActorCritic(
  Actor(
    Sequential(
      (0): Dense(in_dims=hidden_size, out_dims=hidden_size, bias=True)
      (1): ReLU()
      (2): Dense(in_dims=hidden_size, out_dims=action_dims, bias=True)
      (3): Categorical() or MultivariateNormalDiag()
    )
  )
  Critic(
    Sequential(
      (0): Dense(in_dims=hidden_size, out_dims=hidden_size, bias=True)
      (1): ReLU()
      (2): Dense(in_dims=hidden_size, out_dims=1, bias=True)
    )
  )
)
\end{verbatim}

\paragraph{Memoryless}
The memoryless model is implemented as a four-layer MLP with ReLU activations between layers. The architecture is as follows:
\begin{verbatim}
Memoryless(
  Sequential(
    (0): Dense(in_dims=input_dim, out_dims=hidden_size, bias=True)
    (1): ReLU()
    (2): Dense(in_dims=hidden_size, out_dims=hidden_size, bias=True)
    (3): ReLU()
    (4): Dense(in_dims=hidden_size, out_dims=hidden_size, bias=True)
    (5): ReLU()
    (6): Dense(in_dims=hidden_size, out_dims=hidden_size, bias=True)
  )
)
\end{verbatim}
\paragraph{Recurrent Neural Network}
Our recurrent neural network consists of a dense layer with ReLU activation, a GRU cell, and another dense layer. In the Battleship environment, we insert an extra dense layer after the first dense layer (which outputs a vector with twice the latent size for this environment). This additional layer processes the first layer’s output and the hit-or-miss bit.

\begin{verbatim}
RNN(
  Sequential(
    (0): Dense(in_dims=input_dim, out_dims=hidden_size, bias=True)
    (1): ReLU()
    (2): GRU(in_dims=hidden_size, hidden_size=hidden_size)
  )
)
BattleshipRNN(
  Sequential(
    (0): Dense(in_dims=input_dim, out_features=2*hidden_size, bias=True)
    (1): ReLU()
    (2): Dense(in_dims=2*hidden_size, out_features=hidden_size, bias=True)
    (3): ReLU()
    (3): GRU(input_size=hidden_size, hidden_dim=hidden_size)
  )
)
\end{verbatim}

\paragraph{Transformer}
Our transformer model is taken from a JAX implementation of the transformer in library~\citep{hamon2024trxl}. 
\begin{verbatim}
Transformer(
  Sequential(
    (0): Encoder(in_dims=input_dim, out_dims=embed_size, bias=True)
    (1): PositionalEmbedding()
    (2): for i in num_layer:
            Transformer(value, query, positional_embedding, mask)
  )
)
\end{verbatim}

\paragraph{CNN} For environments with visual inputs, we use the following two CNN architectures based on image resolution. For image larger than 20 pixels, we employ a four-layers convolution network defined as follows:
\begin{verbatim}
FullImageCNN(
  Sequential(
    (0): Conv(features=channels, kernel_size=(7, 7), strides=4)
    (1): ReLU()
    (2): Conv(features=num_channels, kernel_size=(5, 5), strides=2)
    (3): ReLU()
    (4): Conv(features=num_channels, kernel_size=(3, 3), strides=2)
    (5): ReLU()
    (6): Conv(features=num_channels, kernel_size=(3, 3), strides=2)
    (7): Flatten()
    (8): ReLU()
    (9): Dense(in_features=flattened_dim, out_features=hidden_size)
    (10): ReLU()
    (11): Dense(in_features=hidden_size, out_features=hidden_size)
  )
)
\end{verbatim}
For image resolution smaller than 20 pixels, we use a three-layers convolutional network with kernel size and strides specific to each domain.
\begin{verbatim}
SmallImageCNN(
  Sequential(
    (0): Conv(features=num_channels, kernel_size, strides)
    (1): ReLU()
    (2): Conv(features=num_channels, kernel_size, strides)
    (3): ReLU()
    (6): Conv(features=num_channels, kernel_size, strides)
    (7): ReLU()
    (8): Flatten()
    (9): Dense(in_features=flattened_dim, out_features=hidden_size)
  )
)
\end{verbatim}

\section{Environment and Hyperparameter details}
\label{appx:env_and_hyper_details}
All our environments are implemented in JAX~\citep{jax2018github} for hardware acceleration. A set of hyperparameters remains constant throughout our experiments. These common settings are provided in Table \ref{tab:common_hyperparams}. Unless otherwise specified, these default parameters were used in every experiment. We also note that unless otherwise stated, the ``fully observable'' agent was trained with a memoryless MLP.
We begin with a discussion on differences observed between our results and other benchmark results in partial observability, then elucidate the full details of each environment.

\subsection{Differences With Other Benchmarks}
\label{appx:differences_benchmarks}
Other works have shown confounding results in terms of benchmarking on partially observable environments. In the POPGym benchmark~\citep{morad2023popgym}, the results imply that memory-based architectures do not help in game environments, such as Battleship. Other works have also shown that in certain cases, memory-based architectures perform worse on fully-observable environments.

These discrepancies arise due to the minute differences in hyperparameters swept and testing methodology. In many of these other works, hyperparameters were not swept, and in many cases kept at the default PPO hyperparameters. This further emphasizes the importance of a fast, GPU-scalable benchmark suite that allows for large-scale hyperparameter sweeps when conducting experiments with partially observable environments.

\begin{table}[ht]
    \centering
    \begin{tabular}{p{4cm} p{3cm} p{8cm}}
        \hline
        \textbf{Hyperparam Name} & \textbf{Value} & \textbf{Description} \\
        \hline
        \texttt{num\_envs} & 4 & number of environments run in parallel \\
        \texttt{default\_max\_steps
        \_in\_episode} & 1000 & maximum steps allowed per episode \\
        \texttt{num\_steps} & 128 & number of steps per update iteration \\
        \texttt{num\_minibatches} & 4 & number of minibatches for gradient updates \\
        \texttt{double\_critic} & False & whether to use $\lambda$-discrepancy \\
        \texttt{action\_concat} & False & whether to concatenate actions with observations \\
        \texttt{lr} & [2.5e-4] & learning rate(s) for the optimizer \\
        \texttt{lambda0} & [0.95] & GAE \(\lambda\) parameter for advantage estimation \\
        \texttt{lambda1} & [0.5] & $\lambda$-discrepancy GAE \(\lambda\) parameter\\
        \texttt{alpha} & [1.0] & weighting factor for combining advantages\\
        \texttt{ld\_weight} & [0.0] & weight in $\lambda$-discrepancy loss\\
        \texttt{vf\_coeff} & [0.5] & value coeffient \\
        \texttt{hidden\_size} & 128 & hidden size of network \\
        \texttt{total\_steps} & \(1.5 \times 10^6\) & total number of training steps \\
        \texttt{entropy\_coeff} & 0.01 & entropy regularization coefficient \\
        \texttt{clip\_eps} & 0.2 & clipping parameter for PPO updates \\
        \texttt{max\_grad\_norm} & 0.5 & maximum gradient norm for clipping\\
        \texttt{anneal\_lr} & True & whether to anneal the learning rate during training \\
        \texttt{image\_size} & 32 & size of input images \\
        \texttt{save\_checkpoints} & False & whether to save checkpoints during training \\
        \texttt{save\_runner\_state} & False & whether to save the final runner state \\
        \texttt{seed} & 2020 & base random seed \\
        \texttt{n\_seeds} & 5 & number of seeds to generate from base random seed\\
        \texttt{qkv\_features} & 256 & feature size for transformer query, key, and value \\
        \texttt{embed\_size} & 256 & embedding size used in the transformer model \\
        \texttt{num\_heads} & 8 & number of attention heads in the transformer \\
        \texttt{num\_layers} & 2 & number of transformer layers \\
        \texttt{window\_mem} & 128 & memory window size for caching hidden states\\
        \texttt{window\_grad} & 64 & gradient window size \\
        \texttt{gating} & True & whether to apply gating in transformer \\
        \texttt{gating\_bias} & 2.0 & bias value for the gating mechanism \\
        \hline
    \end{tabular}
    \caption{Default Hyperparameter Settings}
    \label{tab:common_hyperparams}
\end{table}

\newpage

\subsection{T-Maze}
T-Maze~\citep{bakker2001reinforcement} is a classic memory testing environment. The agent starts off with equal probability in one of two hallways: a hallway where the reward is up, and a hallway where the reward is down. At the first grid, the agent is informed which hallway its in. After leaving the first grid, the observations no longer inform the agent which hallway it is in, and the agent has to remember its initial observations until it reaches the junction. T-Maze 10 is this maze with a hallway length of 10.

\paragraph{Observation Space} The agent's observation is a binary vector with 4 elements. The first two elements dictate which hallway the agent is in (reward up or reward down) and is only set at the start grid. The next element is 1 if the agent is in the hallway. The third elemnt is 1 if the agent is in the junction.

\paragraph{Full Observation Space} The full observation space environment has the same observation shape, but the first two elements are always set according to which hallway the agent is in.

\paragraph{Action Space} The action space is discrete with 4 possible actions, corresponding to moving in the four cardinal direcitons.

\paragraph{Reward} The agent gets +4 for going to the correct side of the junction, and -0.1 for going to the wrong side.

\paragraph{Hyperparameter} For T-Maze 10, we conduct a hyperparameter sweep over 5 seeds for all hyperparameters in Table \ref{table:rnn_tmaze_hyperparams} for memoryless, recurrent PPO, Transformer-XL and fully observable. For LD experiments, we sweep through Table \ref{table:ld_tmaze_hyperparams}. We set the hidden size to 32. We train all algorithms for $1 \times 10^6$ steps and the best hyperparameters are reported in Table \ref{table:tmaze_env_hyperparams}. Then we rerun the experiments over 30 seeds using best hyperparameters.

\begin{table*}[htbp]
    \centering
    \begin{tabular}{l l}
        \hline
        \textbf{Hyperparameter} \\
        \hline
        Step size & 
            \(\{2.5 \times 10^{-3},\, 2.5 \times 10^{-4},\, 2.5 \times 10^{-5},\, 2.5 \times 10^{-6}\}\) \\
        \(\lambda_0\) & 
            \(\{0.1,\, 0.3,\, 0.5,\, 0.7,\, 0.9,\, 0.95\}\) \\
        \hline
    \end{tabular}
    \caption{T-Maze-10 hyperparameters swept across non-Lambda discrepancy algorithms.}
    \label{table:rnn_tmaze_hyperparams}
\end{table*}

\begin{table*}[ht]
    \centering
    \begin{tabular}{l l}
        \hline
        \textbf{Hyperparameter} \\
        \hline
        Step size & 
            \(\{2.5 \times 10^{-3},\, 2.5 \times 10^{-4},\, 2.5 \times 10^{-5}\}\) \\
        \(\lambda_0\) & 
            \(\{0.1,\, 0.5,\, 0.95\}\) \\
        \(\lambda_1\) & 
            \(\{0.5,\, 0.7,\,0.95\}\) \\
        \(\beta\) & 
            \(\{0.25,\, 0.5\}\)\\
        \hline
    \end{tabular}
    \caption{T-Maze-10 hyperparameters swept across Lambda discrepancy algorithm.}
    \label{table:ld_tmaze_hyperparams}
\end{table*}

\begin{table*}[ht]
    \centering
    \begin{tabular}{l c c c c}
        \hline
         & \textbf{Step size} & \boldmath{\(\lambda_0\)} & \boldmath{\(\lambda_1\)} & \boldmath{\(\beta\)} \\
        \hline
        Fully Observable & 
            \(2.5 \times 10^{-4}\) & 
            \(0.5\) & 
            -- & 
            -- \\
        Memoryless & 
            \(2.5 \times 10^{-4}\) & 
            \(0.3\) & 
            -- & 
            -- \\
        RNN & 
            \(2.5 \times 10^{-3}\) & 
            \(0.7\) & 
            -- & 
            -- \\
        Transformer-XL & 
            \(2.5 \times 10^{-4}\) & 
            \(0.9\) & 
            -- & 
            -- \\
        Lambda Discrepancy & 
            \(2.5 \times 10^{-4}\) & 
            \(0.95\) & 
            \(0.95\) & 
            \(0.5\) \\
        \hline
    \end{tabular}
    \caption{T-Maze 10 Best Hyperparameters}
    \label{table:tmaze_env_hyperparams}
\end{table*}

\subsection{Rocksample}
Rocksample \citep{smith2004rocksample} is a navigation problem that simulates a rover searching the environment and assess the rocks. In a rocksample$(n, k)$ problem, $n$ represents the size of the grid and $k$ represents the number of rock in the environments. In our experiments, we consider two variants: Rocksample$(11, 11)$ and Rocksample$(15, 15)$. At the start of each run, rock positions are sampled randomly, and every rock is independently assigned a status of either good or bad. The goal of the agent is to sample all the good rock and avoid all the back ones.

\paragraph{Observation Space} The agent's observation is a binary vector with $2n+k$ elements. The first $2n$ elements encode the agent's positions on the board using a two-hot representation. The remaining k elements are only updated after the agent either checks or samples a rock and the corresponding i elements is set to 1 if ith rock appear to be good.

\paragraph{Full Observation Space} The full observation space of RockSample is a ``perfect memory'' state representation, also with $2n+k$ elements. The first $2n$ elements are the same positional encoding. The final $k$ elements keep the most recent observation seen from each $k$ rocks, either from checking or sampling a rock.

\paragraph{Action Space} The action space is $(5+k, )$. The first four dimensions correspond to movement of the agent. The fifth dimension corresponds to sampling a rock in its current position. The last k dimensions correspond to checking each rock. When the agent checks a rock, it receives the rock's correct parity with probability determined by the half-efficiency distance, which is based on the distance from the rock being checked:
\begin{equation}
    \frac{1}{2} \left( 1 + 2^{-d / max_d} \right),
\end{equation}
where $d$ is the $l_2$ distance to the rock, and $max_d$ is the maximum distance from any grid in the domain. This means the closer an agent is to a rock, the more likely the agent will get the correct parity. 

\paragraph{Reward} The agent gets +10 for exiting to the east. The agent also gets +10 for sampling a good rock, and -10 for sampling a bad rock.

\paragraph{Hyperparameter} For both Rocksample(11,11) and Rocksample(15,15), we conduct a hyperparameter sweep over 5 seeds for all hyperparameters in Table \ref{table:rnn_tmaze_hyperparams} for memoryless, recurrent PPO, Transformer-XL and fully observable. For LD experiments, we sweep through Table \ref{table:ld_tmaze_hyperparams}. In Rocksample(11,11), we set the hidden size to 256, the number of environments to 8 and entropy coefficient to 0.2. In Rocksample(15,15), we set the hidden size to 512, number of environments to 16 and entropy coefficient to 0.2. We train all algorithms for $5 \times 10^6$ steps and the best hyperparameters are reported in Table \ref{table:rnn_tmaze_hyperparams} and Table \ref{table:ld_tmaze_hyperparams}. Then we rerun the experiments over 30 seeds using best hyperparameters. For both Rocksample(11, 11) and Rocksample(15, 15), the perfect memory agent was trained with an RNN as opposed to a memoryless MLP. This was due to improved function approximation by the RNN, even with a fully observable state.

% \begin{table*}[htbp]
%     \centering
%     \begin{tabular}{l l}
%         \hline
%         \textbf{Hyperparameter} \\
%         \hline
%         Step size & 
%             \(\{2.5 \times 10^{-3},\, 2.5 \times 10^{-4},\, 2.5 \times 10^{-5},\, 2.5 \times 10^{-6}\}\) \\
%         \(\lambda_0\) & 
%             \(\{0.1,\, 0.3,\, 0.5,\, 0.7,\, 0.9,\, 0.95\}\) \\
%         \hline
%     \end{tabular}
%     \caption{Rocksample hyperparameters swept across non-Lambda discrepancy algorithms.}
%     \label{table:rnn_rs_hyperparams}
% \end{table*}

% \begin{table*}[ht]
%     \centering
%     \begin{tabular}{l l}
%         \hline
%         \textbf{Hyperparameter} \\
%         \hline
%         Step size & 
%             \(\{2.5 \times 10^{-3},\, 2.5 \times 10^{-4},\, 2.5 \times 10^{-5}\}\) \\
%         \(\lambda_0\) & 
%             \(\{0.1,\, 0.5,\, 0.95\}\) \\
%         \(\lambda_1\) & 
%             \(\{0.5,\, 0.7,\,0.95\}\) \\
%         \(\beta\) & 
%             \(\{0.25,\, 0.5\}\)\\
%         \hline
%     \end{tabular}
%     \caption{Rocksample hyperparameters swept across Lambda discrepancy algorithm.}
%     \label{table:ld_rs_hyperparams}
% \end{table*}

\begin{table*}[ht]
    \centering
    \begin{tabular}{l c c c c}
        \hline
         & \textbf{Step size} & \boldmath{\(\lambda_0\)} & \boldmath{\(\lambda_1\)} & \boldmath{\(\beta\)} \\
        \hline
        Fully Observable & 
            \(2.5 \times 10^{-4}\) & 
            \(0.5\) & 
            -- & 
            -- \\
        Memoryless & 
            \(2.5 \times 10^{-3}\) & 
            \(0.3\) & 
            -- & 
            -- \\
        RNN & 
            \(2.5 \times 10^{-4}\) & 
            \(0.95\) & 
            -- & 
            -- \\
        Transformer-XL & 
            \(2.5 \times 10^{-4}\) & 
            \(0.1\) & 
            -- & 
            -- \\
        Lambda Discrepancy & 
            \(2.5 \times 10^{-3}\) & 
            \(0.1\) & 
            \(0.95\) & 
            \(0.25\) \\
        \hline
    \end{tabular}
    \caption{Rocksample(11, 11) Best Hyperparameters}
    \label{table:rs_11_env_hyperparams}
\end{table*}

\begin{table*}[ht]
    \centering
    \begin{tabular}{l c c c c}
        \hline
         & \textbf{Step size} & \boldmath{\(\lambda_0\)} & \boldmath{\(\lambda_1\)} & \boldmath{\(\beta\)} \\
        \hline
        Fully Observable & 
            \(2.5 \times 10^{-4}\) & 
            \(0.7\) & 
            -- & 
            -- \\
        Memoryless & 
            \(2.5 \times 10^{-3}\) & 
            \(0.3\) & 
            -- & 
            -- \\
        RNN & 
            \(2.5 \times 10^{-4}\) & 
            \(0.95\) & 
            -- & 
            -- \\
        Transformer-XL & 
            \(2.5 \times 10^{-5}\) & 
            \(0.3\) & 
            -- & 
            -- \\
        Lambda Discrepancy & 
            \(2.5 \times 10^{-4}\) & 
            \(0.5\) & 
            \(0.5\) & 
            \(0.25\) \\
        \hline
    \end{tabular}
    \caption{Rocksample(15, 15) Best Hyperparameters}
    \label{table:rs_15_env_hyperparams}
\end{table*}

\subsection{Battleship}
Partially observable battleship~\citep{silver2010pomcp} is a less observable variant of the traditional battleship game. The agent has a $10\times10$ board and four ships with length \{5, 4, 3, 2\} that are uniformly random generated on the board at the start of an episode. The agent's objective is to hit all parts of each ship under the condition that no position was allowed to hit twice. This setup results in a finite horizon problem, with a maximum of 100 moves (one for each grid position). Therefore, we set the discounted factor  $\gamma = 1$. The environment terminates when all positions on the grid with a ship are hit.

\paragraph{Observation Space} After each step, the agent only receives a single binary signal. A 0 indicate no ship is hit and a 1 indicate the opposite. To simplify the learning process, we concatenate the agent's last action to the observation. Since the action size is $10 \times 10$. The observation space is (101, )

\paragraph{Action Space} The action space is defined as $\{1, \dots, 10\} \times \{1, \dots, 10\}$, which correspond to row and column number of the board that indicate the next target to hit. Actions are masked at each step to prevent illegal moves.

\paragraph{Reward} The agent is penalised $-1$ for every step it took. When all ships are hit, the agent receive a reward of $100$.

\paragraph{Hyperparameter} We conducted a hyperparameter sweep over 10 seeds across memoryless, fully observable, RNN, and Transformer-XL models using all the parameters in Table \ref{table:rnn_tmaze_hyperparams}, and swept the hyperparameters in Table \ref{table:ld_tmaze_hyperparams} for LD. All experiments are trained for $1\times10^7$ steps to select the best hyperparameters. The entropy coefficient was adjusted to 0.05 to encourage exploration, the hidden size was set to 512, and the number of environments was set to 32. Additionally, We set steps-log-frequency to 8 and update-log-frequency to 10. The best hyperparameters selected after the sweep are summarized in Table \ref{table:battleship_best_hyperparam}. Then we rerun the experiments over 30 seeds using best hyperparameters.

\begin{table*}[ht]
    \centering
    \begin{tabular}{l c c c c}
        \hline
          & \textbf{Step size} & \boldmath{\(\lambda_0\)} & \boldmath{\(\lambda_1\)} & \boldmath{\(\beta\)} \\
        \hline
        Fully Observable & 
            -- & 
            -- & 
            -- & 
            -- \\
        Memoryless & 
            \(2.5 \times 10^{-3}\) & 
            \(0.1\) & 
            -- & 
            -- \\
        RNN & 
            \(2.5 \times 10^{-3}\) & 
            \(0.7\) & 
            -- & 
            -- \\
        Transformer-XL & 
            \(2.5 \times 10^{-5}\) & 
            \(0.1\) & 
            -- & 
            -- \\
        Lambda Discrepancy & 
            \(2.5 \times 10^{-3}\) & 
            \(0.1\) & 
            \(0.95\) & 
            \(0.5\) \\
        \hline
    \end{tabular}
    \caption{Battleship Best Hyperparameters}
    \label{table:battleship_best_hyperparam}

\end{table*}

\subsection{Masked Continuous Control} 
\label{appx:masked_mujoco}

Masked continuous control are Mujoco environments~\citep{todorov2012mujoco,brax2021github} with only velocity (Vel. Only) or only positional (Pos. Only) features. 
\paragraph{Observation Space} 
The observation space for each environment changes depending on which environment is used and what variables are masked. We refer to our code repository (\repoURL) for full details of each observation space, as well as the Brax documentation~\citep{brax2021github} for details of the original observation space. Note that all masked continuous control results presented in this work was smoothed using a Savitzky-Golay filter~\citep{savitzky1964savgolfilter} with a window of $30$ and a polynomial degree of $3$.

\paragraph{Fully Observable Observation Space} The full observation of each environment are equivalent to the full observations in each Brax environment.

\paragraph{Reward and Action Space} The reward and action space are similar to the corresponding Brax environment.

\paragraph{Hyperparameter} For all environments, we conduct a hyperparam sweep over 5 seeds for all hyperparameters in Table \ref{table:rnn_tmaze_hyperparams}, Table \ref{table:ld_tmaze_hyperparams}. We trained for $5 \times 10^7$ steps. The hidden size is set to 256, step-log-frequency to 16, update-log-frequency to 20. For transformer, the embed size is set to 96. We list the best hyperparameters for the Walker-V and HalfCheetah-V environments in Tables~\ref{table:halfcheetah_v_hyperparams} and~\ref{table:walker_v_hyperparams} as they appear in our benchmark. 
We refer to our codebase for the best hyperparameters selected for the full masked mujoco hyperparameter sweep. For both of these environments, we use RNN function approximation for the fully observable results, due to better performance. This is different from the fully-observable agents run in the experiments run on the full set of masked Mujoco environments, where the fully observable version used a memoryless fully connected neural network.

\begin{table*}[ht]
    \centering
    \caption{Halfcheetah-V Best Hyperparameters}
    \label{table:halfcheetah_v_hyperparams}
    \begin{tabular}{l c c c c}
        \hline
          & \textbf{Step size} & \boldmath{\(\lambda_0\)} & \boldmath{\(\lambda_1\)} & \boldmath{\(\beta\)} \\
        \hline
        Fully Observable & 
            \(2.5 \times 10^{-4}\) & 
            \(0.1\) & 
            -- & 
            -- \\
        Memoryless & 
            \(2.5 \times 10^{-4}\) & 
            \(0.7\) & 
            -- & 
            -- \\
        RNN & 
            \(2.5 \times 10^{-4}\) & 
            \(0.9\) & 
            -- & 
            -- \\
        Transformer-XL & 
            \(2.5 \times 10^{-4}\) & 
            \(0.9\) & 
            -- & 
            -- \\
        Lambda Discrepancy & 
            \(2.5 \times 10^{-5}\) & 
            \(0.95\) & 
            \(0.7\) & 
            \(0.25\) \\
        \hline
    \end{tabular}
\end{table*}
\begin{table*}[ht]
    \centering
    \caption{Walker-V Best Hyperparameters}
    \label{table:walker_v_hyperparams}
    \begin{tabular}{l c c c c}
        \hline
          & \textbf{Step size} & \boldmath{\(\lambda_0\)} & \boldmath{\(\lambda_1\)} & \boldmath{\(\beta\)} \\
        \hline
        Fully Observable & 
            \(2.5 \times 10^{-4}\) & 
            \(0.9\) & 
            -- & 
            -- \\
        Memoryless & 
            \(2.5 \times 10^{-4}\) & 
            \(0.95\) & 
            -- & 
            -- \\
        RNN & 
            \(2.5 \times 10^{-4}\) & 
            \(0.95\) & 
            -- & 
            -- \\
        Transformer-XL & 
            \(2.5 \times 10^{-5}\) & 
            \(0.95\) & 
            -- & 
            -- \\
        Lambda Discrepancy & 
            \(2.5 \times 10^{-4}\) & 
            \(0.95\) & 
            \(0.95\) & 
            \(0.5\) \\
        \hline
    \end{tabular}
\end{table*}

\subsection{DeepMind Lab MiniGrid mazes}
\label{appx:navix}

\begin{figure}[t]
    \centering
    % \vspace{-20px}
    \begin{subfigure}[b]{0.38\linewidth}
        \begin{minipage}{\linewidth}
            \centering
            \includegraphics[width=\linewidth]{images/navix_01.pdf}
        \end{minipage}
    \end{subfigure}
    \hfill
    \begin{subfigure}[b]{0.58\linewidth}
        % \vspace{22px}
        \begin{minipage}{\linewidth}
            \centering
            \includegraphics[width=\linewidth]{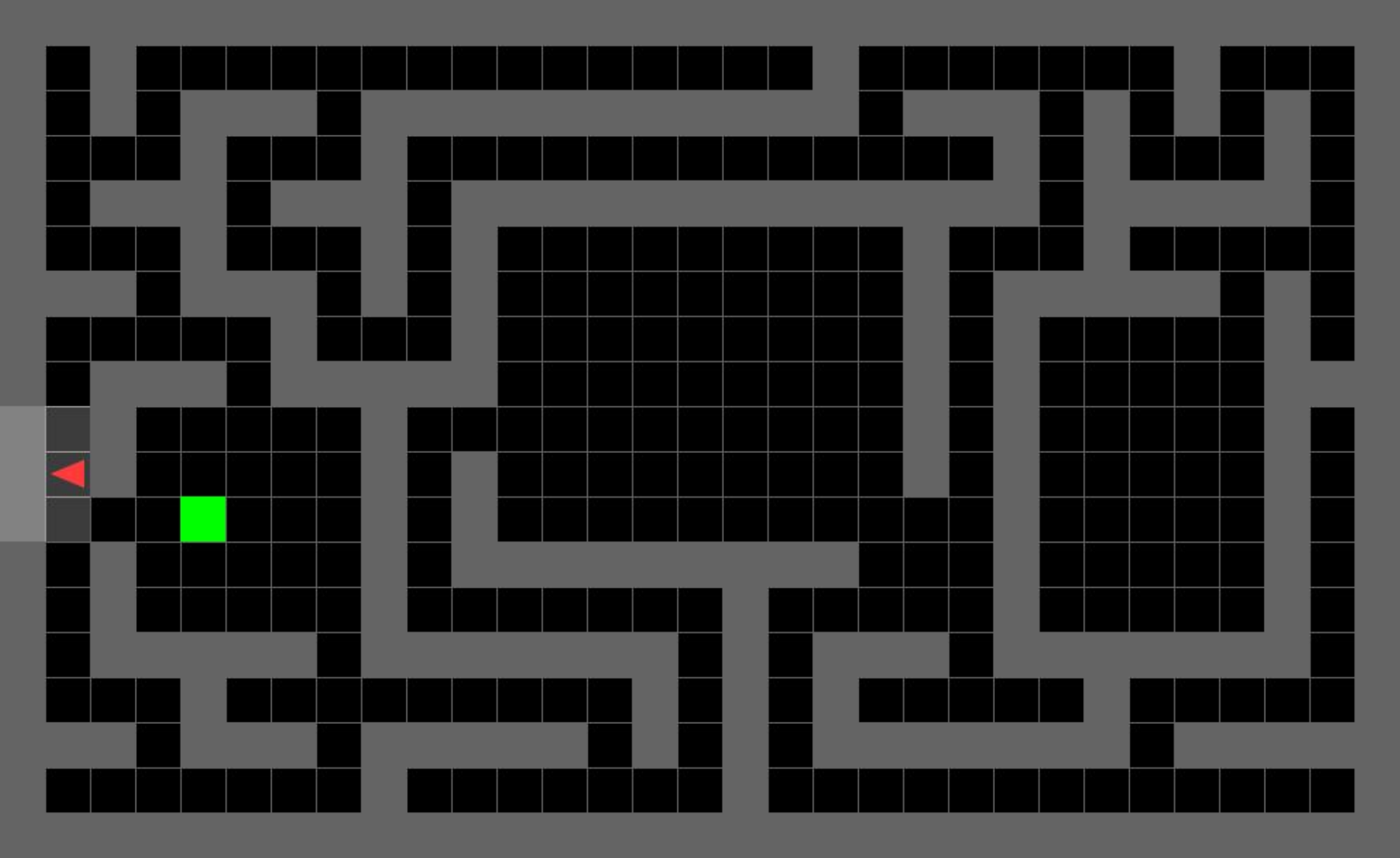}
        \end{minipage}
    \end{subfigure}
    \hfill
    \begin{subfigure}[b]{0.7\linewidth}
        % \vspace{22px}
        \begin{minipage}{\linewidth}
            \centering
            \includegraphics[width=\linewidth]{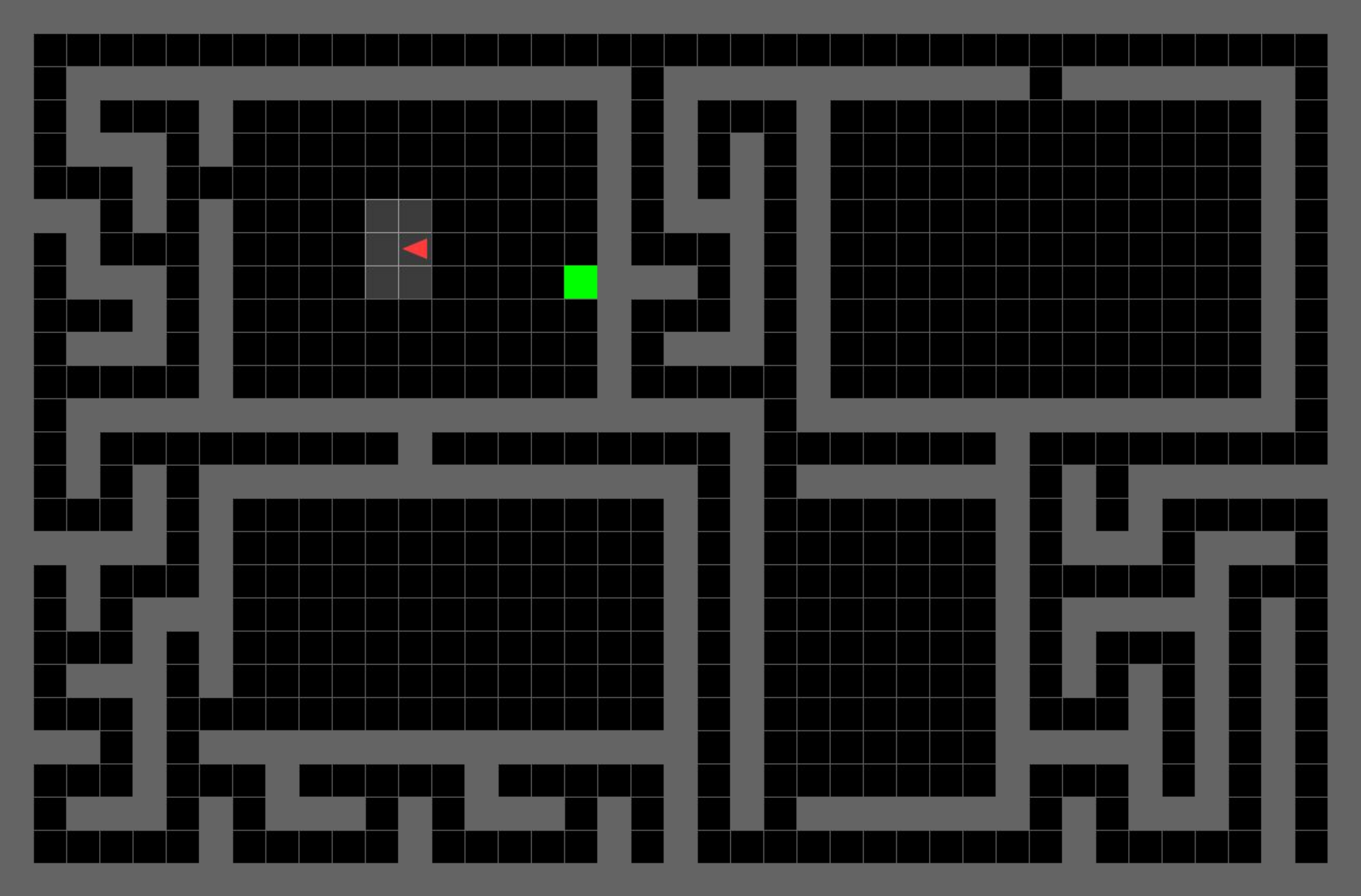}
        \end{minipage}
    \end{subfigure}
    \caption{(Left to right, top to down) Three DeepMind Lab MiniGrid mazes, \texttt{maze\_id = 01, 02, 03}. As \texttt{maze\_id} increases, maze complexity and size increases as well.}
    \label{fig:dmlab_mazes}
\end{figure}
DeepMind Lab MiniGrid mazes are MiniGrid~\citep{boisvert2023minigrid, pignatelli2024navix} mazes with the maze layouts from the DeepMind Lab~\citep{beattie2016dmlab} ``navigation levels with a static map layout'' as shown in Figure~\ref{fig:dmlab_mazes}. These three mazes get increasingly complex and large. At the beginning of every episode, both agent start state and goal state are randomly initialized. Maximum number of episode steps is 2000, 4000 and 6000 for each maze, from lowest ID to highest ID.

\paragraph{Observation Space} One-hot first-person images of size $(2, 3, 2)$, where the two channels represent the wall positions and goal locations in the $2 \times 3$ grids in front of the agent.

\paragraph{Fully Observable Observation Space} Agent-centric one-hot images of size $(2h -1, 2w - 1, 2 + 4)$, where $h$ and $w$ are the height and width of each maze. Position is encoded by shifting the map so that the agent is always in the center. The first two channels represent the walls and goal positions. The last four dimensions represent a one-hot encoding (across the channels) of the direction the agent is facing.

\paragraph{Action Space} Discrete space of 3 actions, representing \texttt{forward}, \texttt{turn left} and \texttt{turn right}.

\paragraph{Reward} The agent gets $+1$ once it reaches the goal, with a discount factor of $\gamma = 0.99$.

\paragraph{Hyperparameter} For both Navix-01 and Navix-02, we conducted our experiments over 5 seeds for all hyperparameters in Table \ref{table:navix_hyperparams}, \ref{table:navix_LD_hyperparams}. The hidden size is set to 512 and the embed size for transformer experiment is set to 220. The number of environment is set to 256 in Navix-01 and 512 in Navix-02. Navix-01 is trained for $1 \times 10^7$ steps and Navix-02 is trained for $1 \times 10^8$ steps. The best hyperparameters are provided in Table \ref{table:navix_01_hyperparams}, \ref{table:navix_02_hyperparams}.

\begin{table*}[ht]
    \centering
    \begin{tabular}{l l}
        \hline
        \textbf{Hyperparameter} \\
        \hline
        Step size & 
            \(\{2.5 \times 10^{-3},\, 2.5 \times 10^{-4},\, 2.5 \times 10^{-5},\, 2.5 \times 10^{-6}\}\) \\
        \(\lambda_0\) & 
            \(\{0.1,\, 0.5,\, 0.7,\, 0.9,\, 0.95\}\) \\
        \hline
    \end{tabular}
    \caption{DeepMind Lab MiniGrid Maze hyperparameters swept across non-Lambda discrepancy algorithms.}
    \label{table:navix_hyperparams}
\end{table*}
\begin{table*}[ht]
    \centering
    \begin{tabular}{l l}
        \hline
        \textbf{Hyperparameter} \\
        \hline
        Step size & 
            \(\{2.5 \times 10^{-4},\, 2.5 \times 10^{-5}\}\) \\
        \(\lambda_0\) & 
            \(\{0.1,\, 0.95\}\) \\
        \(\lambda_1\) & 
            \(\{0.5,\, 0.7,\,0.95\}\) \\
        \(\beta\) & 
            \(\{0.25,\, 0.5\}\)\\
        \hline
    \end{tabular}
    \caption{DeepMind Lab MiniGrid Maze hyperparameters swept across Lambda discrepancy algorithm.}
    \label{table:navix_LD_hyperparams}
\end{table*}

\begin{table*}[ht]
    \centering
    \begin{tabular}{l c c c c}
        \hline
          & \textbf{Step size} & \boldmath{\(\lambda_0\)} & \boldmath{\(\lambda_1\)} & \boldmath{\(\beta\)} \\
        \hline
        Fully Observable & 
            \(2.5 \times 10^{-4}\) & 
            \(0.95\) & 
            -- & 
            -- \\
        Memoryless & 
            \(2.5 \times 10^{-4}\) & 
            \(0.95\) & 
            -- & 
            -- \\
        RNN & 
            \(2.5 \times 10^{-4}\) & 
            \(0.9\) & 
            -- & 
            -- \\
        Transformer-XL & 
            \(2.5 \times 10^{-4}\) & 
            \(0.95\) & 
            -- & 
            -- \\
        Lambda Discrepancy & 
            \(2.5 \times 10^{-4}\) & 
            \(0.95\) & 
            \(0.5\) & 
            \(0.25\) \\
        \hline
    \end{tabular}
    \caption{DeepMind Lab MiniGrid Maze Level 1 Best Hyperparameters}
    \label{table:navix_01_hyperparams}
\end{table*}
\begin{table*}[ht]
    \centering
    \begin{tabular}{l c c c c}
        \hline
          & \textbf{Step size} & \boldmath{\(\lambda_0\)} & \boldmath{\(\lambda_1\)} & \boldmath{\(\beta\)} \\
        \hline
        Fully Observable & 
            \(2.5 \times 10^{-4}\) & 
            \(0.95\) & 
            -- & 
            -- \\
        Memoryless & 
            \(2.5 \times 10^{-3}\) & 
            \(0.95\) & 
            -- & 
            -- \\
        RNN & 
            \(2.5 \times 10^{-4}\) & 
            \(0.95\) & 
            -- & 
            -- \\
        Transformer-XL & 
            \(2.5 \times 10^{-4}\) & 
            \(0.95\) & 
            -- & 
            -- \\
        Lambda Discrepancy & 
            \(2.5 \times 10^{-4}\) & 
            \(0.95\) & 
            \(0.95\) & 
            \(0.25\) \\
        \hline
    \end{tabular}
    \caption{DeepMind Lab MiniGrid Maze Level 2 Best Hyperparameters}
    \label{table:navix_02_hyperparams}
\end{table*}

\subsection{Visual Continuous Control}
Visual continuous control are Mujoco environments with pixel features. We integrate the Madrona MJX \citep{Shacklett2024madrona} renderer on top of Brax environments to enable just-in-time (JIT) compilation over rendering in JAX. Note that the Madrona MJX renderer supports only a single batched environment. Thus, we remove the parallelization of training hyperparameters in our algorithms specifically for visual mujoco experiment. Also note that all visual continuous control results presented in this work were also smoothed with a Savitzky-Golay filter~\citep{savitzky1964savgolfilter} with a window of $30$ and a polynomial degree of $3$.

\paragraph{Observation Space} To ensure that the environment is memory-improvable, we do not use frame stacking. The observation space is a single frame represent the current view of the agent. Height and width of the image are determined by the image size hyperparameter. In our experiments, we set the image size to 32, so the observation is (32, 32, 3) for visual mujoco experiments.

\paragraph{Fully Observable Observation Space} For fully observable ceiling, we use the observation space from original Brax \citep{brax2021github} environments for the HalfCheetah and Ant environments. Halfcheetah has observation space (18, ) and Ant has observation space (27, ). 
% See the details in Table~\ref{table:visual_mujoco_fully_observable}.

\paragraph{Action Space} We use Brax HalfCheetah and Ant action space to evaluate all our algorithm. HalfCheetah has a continuous action space of shape (6, ) and Ant has a continuous action space of shape (8, ). The values of actions in both of the environments fall between -1 and 1, where each component representing the torque applied to a specific part of the agent.

\paragraph{Reward} The reward function of Ant consists three parts. The agent is rewarded for every second it survives and It is also rewarded for moving in the desired direction. It is penalised for taking too large action and also if the external force is too large. The reward function of Halfcheetah has two parts. The agent is rewarded for going in forward direction and it is penalised for taking too large action.

\paragraph{Hyperparameter} We swept both Halfcheetah and Ant over 3 seeds for all hyperparameter in Table \ref{table:navix_hyperparams}, \ref{table:navix_LD_hyperparams} and train for $5 \times 10^6$ to get the best hyperparameters. Specifically, we set the hidden size to 512. For transformer experiments, we set the embed size to 220 to match the total number of parameters in recurrent PPO. The rest hyperparameters are default.  We present the best hyperparameters found for environments in Table \ref{table:visual_ant_best_hyperparams}, \ref{table:visual_halfcheetah_best_hyperparams}. After the selection, we rerun the experiments over 30 seeds using best hyperparameters.

\begin{table}[ht]
    \centering
    \begin{tabular}{l c c c c}
        \hline
          & \textbf{Step size} & \boldmath{\(\lambda_0\)} & \boldmath{\(\lambda_1\)} & \boldmath{\(\beta\)} \\
        \hline
        Fully Observable & 
            \(2.5 \times 10^{-4}\) & 
            \(0.5\) & 
            -- & 
            -- \\
        Memoryless & 
            \(2.5 \times 10^{-4}\) & 
            \(0.1\) & 
            -- & 
            -- \\
        RNN & 
            \(2.5 \times 10^{-4}\) & 
            \(0.7\) & 
            -- & 
            -- \\
        Transformer-XL & 
            \(2.5 \times 10^{-4}\) & 
            \(0.5\) & 
            -- & 
            -- \\
        Lambda Discrepancy & 
            \(2.5 \times 10^{-4}\) & 
            \(0.1\) & 
            \(0.5\) & 
            \(0.5\) \\
        \hline
    \end{tabular}
    \caption{Ant Best Hyperparameters}
    \label{table:visual_ant_best_hyperparams}
\end{table}

\begin{table}[ht]
    \centering
    \begin{tabular}{l c c c c}
        \hline
          & \textbf{Step size} & \boldmath{\(\lambda_0\)} & \boldmath{\(\lambda_1\)} & \boldmath{\(\beta\)} \\
        \hline
        Fully Observable & 
            \(2.5 \times 10^{-4}\) & 
            \(0.5\) & 
            -- & 
            -- \\
        Memoryless & 
            \(2.5 \times 10^{-4}\) & 
            \(0.7\) & 
            -- & 
            -- \\
        RNN & 
            \(2.5 \times 10^{-4}\) & 
            \(0.7\) & 
            -- & 
            -- \\
        Transformer-XL & 
            \(2.5 \times 10^{-4}\) & 
            \(0.7\) & 
            -- & 
            -- \\
        Lambda Discrepancy & 
            \(2.5 \times 10^{-4}\) & 
            \(0.95\) & 
            \(0.95\) & 
            \(0.5\) \\
        \hline
    \end{tabular}
    \caption{Halfcheetah Best Hyperparameters}
    \label{table:visual_halfcheetah_best_hyperparams}
\end{table}

\subsection{No-inventory Crafter}
No-inventory Crafter is a more partially observable variant of Crafter \citep{hafner2021crafter}. This environments was built on top of the Craftax framework \citep{matthews2024craftax}. Craftax is a version of Crafter that is implemented in JAX~\citep{jax2018github}. On top of their work, we furthur made this environment more partially observable by masking the inventory located at the bottom of an observation. 

\paragraph{Observation Space} The original Craftax observation consists of a grid of 13 by 9 pixel squares, where each square is $10 \times 10$ pixels, making the original observation $(130, 90, 3)$. To make it more efficient, our No-inventory Crafter pixel observation has the same form as Craftax, but we downscaled the square from $10 \times 10$ to $3 \times 3$ and then we mask the pixels that correspond to the inventory, resulting in a final observation shape $(27, 33, 3)$.

\paragraph{Fully Observable Observation Space} For our fully observable ceiling, we use the Craftax symbolic observation, which has shape (8268, ). The first section is the flattened map representation containing information about block, item, mob and light level. Then the next section is the inventory, followed by potions, player's intrinsics, player's direction, armour and special values. 

\paragraph{Action Space} We use the same action space with Craftax, which is a discrete action space of 43. Note that every action can be taken at any time, thus attempting to execute an action that is not available will result in a no-op action. 

\paragraph{Reward} We adopt the same reward scheme used in Craftax. The agent receive the reward the first time it complete an achievement. There are a total 65 achievements which are characterized into 4 categories: ‘Basic’, ‘Intermediate’, ‘Advanced’, and ‘Very Advanced’, for which the agent is rewarded 1, 3, 5, 8 points respectively. The agent is also penalised 0.1 point every point of damage it took and rewarded 0.1 every health it recovered.

\paragraph{Hyperparameter} We swept Craftax over 3 seeds for all hyperparameters in Table \ref{table:navix_hyperparams} and \ref{table:navix_LD_hyperparams} and train for $5 \times 10^8$ steps across all the algorithms. We set the number of environments to 256 and hidden size to 512. For the transformer experiments, we set the embed size to 220 to match the total number of parameters in recurrent PPO. The best hyperparameters selected after the sweep are summarized in Table~\ref{table:craftax_hyperparams}. After selection, we rerun the experiments over 30 seeds with the best hyperparameters.

\begin{table}[ht]
    \centering
    \caption{No-inventory Crafter Best Hyperparameters}
    \label{table:craftax_hyperparams}
    \begin{tabular}{l c c c c}
        \hline
          & \textbf{Step size} & \boldmath{\(\lambda_0\)} & \boldmath{\(\lambda_1\)} & \boldmath{\(\beta\)} \\
        \hline
        Fully Observable & 
            \(2.5 \times 10^{-4}\) & 
            \(0.7\) & 
            -- & 
            -- \\
        Memoryless & 
            \(2.5 \times 10^{-5}\) & 
            \(0.95\) & 
            -- & 
            -- \\
        RNN & 
            \(2.5 \times 10^{-4}\) & 
            \(0.5\) & 
            -- & 
            -- \\
        Transformer-XL & 
            \(2.5 \times 10^{-5}\) & 
            \(0.7\) & 
            -- & 
            -- \\
        Lambda Discrepancy & 
            \(2.5 \times 10^{-4}\) & 
            \(0.1\) & 
            \(0.95\) & 
            \(0.25\) \\
        \hline
    \end{tabular}
\end{table}

\subsection{Ablation studies} 
\label{appx:ablations}

Here we describe details of the number of parallel environments and network width ablation studies. Both studies were conducted over $5$ seeds. Hyperparameters for the ablation study on number of parallel environments swept were the same as Appendix~\ref{appx:navix} for \texttt{maze\_id = 01}, except with the additional sweep of \texttt{num\_envs}$\in (64, 256)$.
Hyperparameters for the ablation study on network width were the same as Appendix~\ref{appx:masked_mujoco}, except with the additional sweep of \texttt{hidden\_size}$\in (32, 64, 256)$.
Best performance was taken over discounted returns.

\end{document}